\documentclass[10pt,twocolumn,letterpaper]{article}

\usepackage[pagenumbers]{cvpr}              %

\definecolor{cvprblue}{rgb}{0.21,0.49,0.74}
\usepackage[pagebackref,breaklinks,colorlinks,allcolors=cvprblue]{hyperref}
\usepackage{graphicx}
\usepackage{amsmath}
\usepackage{amssymb}
\usepackage{tabularx,booktabs}
\newcolumntype{Y}{>{\centering\arraybackslash}X} %

\usepackage{cuted} %
\usepackage{xcolor}

\usepackage{array}
\usepackage{subcaption}
\usepackage{stfloats}
\usepackage{placeins}

\usepackage[pagebackref,breaklinks,colorlinks]{hyperref}
\usepackage[capitalize]{cleveref}
\usepackage{balance}

\usepackage{comment}
\usepackage{amsmath,amssymb} %
\usepackage{wrapfig} %
\usepackage{color}

\usepackage{mathtools}
\usepackage[accsupp]{axessibility}  %

\usepackage[utf8]{inputenc} %
\usepackage[T1]{fontenc}    %
\usepackage{xr-hyper}
\usepackage{url}            %
\usepackage{booktabs}       %
\usepackage{xcolor,colortbl}         %
\usepackage{caption}

\usepackage{multirow}
\usepackage{enumitem}

\usepackage[normalem]{ulem}

\usepackage{subcaption}
\usepackage{float}
\usepackage{lscape}     

\usepackage{xspace}
\usepackage{setspace}
\usepackage{pifont}
\usepackage{comment}

\usepackage{enumitem}%

\def\paperID{} %
\def\confName{CVPR}
\def\confYear{2026}

\title{Re-Depth Anything: Test-Time Depth Refinement via Self-Supervised Re-lighting}

\author{Ananta R. Bhattarai\\
Bielefeld University\\
\and
Helge Rhodin\\
Bielefeld University\\
}

\begin{document}
\newcommand{\parag}[1]{\noindent\textbf{#1}} %

\newcommand{\R}{\mathbb{R}}

\newcommand{\va}{\mathbf{a}}
\newcommand{\vb}{\mathbf{b}}
\newcommand{\vc}{\mathbf{c}}
\newcommand{\vd}{\mathbf{d}}
\newcommand{\ve}{\mathbf{e}}
\newcommand{\vf}{\mathbf{f}}
\newcommand{\vg}{\mathbf{g}}
\newcommand{\vh}{\mathbf{h}}
\newcommand{\vi}{\mathbf{i}}
\newcommand{\vj}{\mathbf{j}}
\newcommand{\vk}{\mathbf{k}}
\newcommand{\vl}{\mathbf{l}}
\newcommand{\vm}{\mathbf{m}}
\newcommand{\vn}{\mathbf{n}}
\newcommand{\vo}{\mathbf{o}}
\newcommand{\vp}{\mathbf{p}}
\newcommand{\vq}{\mathbf{q}}
\newcommand{\vr}{\mathbf{r}}
\newcommand{\vt}{\mathbf{t}}
\newcommand{\vu}{\mathbf{u}}
\newcommand{\vv}{\mathbf{v}}
\newcommand{\vw}{\mathbf{w}}
\newcommand{\vx}{\mathbf{x}}
\newcommand{\vy}{\mathbf{y}}
\newcommand{\vz}{\mathbf{z}}

\newcommand{\mA}{\mathbf{A}}
\newcommand{\mB}{\mathbf{B}}
\newcommand{\mC}{\mathbf{C}}
\newcommand{\mD}{\mathbf{D}}
\newcommand{\mE}{\mathbf{E}}
\newcommand{\mF}{\mathbf{F}}
\newcommand{\mG}{\mathbf{G}}
\newcommand{\mH}{\mathbf{H}}
\newcommand{\mI}{\mathbf{I}}
\newcommand{\mJ}{\mathbf{J}}
\newcommand{\mK}{\mathbf{K}}
\newcommand{\mL}{\mathbf{L}}
\newcommand{\mM}{\mathbf{M}}
\newcommand{\mN}{\mathbf{N}}
\newcommand{\mO}{\mathbf{O}}
\newcommand{\mP}{\mathbf{P}}
\newcommand{\mQ}{\mathbf{Q}}
\newcommand{\mR}{\mathbf{R}}
\newcommand{\mS}{\mathbf{S}}
\newcommand{\mT}{\mathbf{T}}
\newcommand{\mU}{\mathbf{U}}
\newcommand{\mV}{\mathbf{V}}
\newcommand{\mW}{\mathbf{W}}
\newcommand{\mX}{\mathbf{X}}
\newcommand{\mY}{\mathbf{Y}}
\newcommand{\mZ}{\mathbf{Z}}

\newcommand{\cA}{\mathcal A}
\newcommand{\cB}{\mathcal B}
\newcommand{\cC}{\mathcal C}
\newcommand{\cD}{\mathcal D}
\newcommand{\cE}{\mathcal E}
\newcommand{\cF}{\mathcal F}
\newcommand{\cG}{\mathcal G}
\newcommand{\cH}{\mathcal H}
\newcommand{\cI}{\mathcal I}
\newcommand{\cJ}{\mathcal J}
\newcommand{\cK}{\mathcal K}
\newcommand{\cL}{\mathcal L}
\newcommand{\cM}{\mathcal M}
\newcommand{\cN}{\mathcal N}
\newcommand{\cO}{\mathcal O}
\newcommand{\cP}{\mathcal P}
\newcommand{\cQ}{\mathcal Q}
\newcommand{\cR}{\mathcal R}
\newcommand{\cS}{\mathcal S}
\newcommand{\cT}{\mathcal T}
\newcommand{\cU}{\mathcal U}
\newcommand{\cV}{\mathcal V}
\newcommand{\cW}{\mathcal W}
\newcommand{\cX}{\mathcal X}
\newcommand{\cY}{\mathcal Y}
\newcommand{\cZ}{\mathcal Z}

\newcommand{\bR}{\mathbb{R}}
\newcommand{\mx}{\mathbf{x}}
\newcommand{\mj}{\mathbf{j}}
\newcommand{\mb}{\mathbf{b}}
\newcommand{\vmu}{\mathbf{\mu}}

\newcommand\blfootnote[1]{%
  \begingroup
  \renewcommand\thefootnote{}\footnote{#1}%
  \addtocounter{footnote}{-1}%
  \endgroup
}

\twocolumn[{
\maketitle
\centering
\includegraphics[width=0.95\textwidth,trim={0, 0.5cm, 0, 0.2cm},clip]{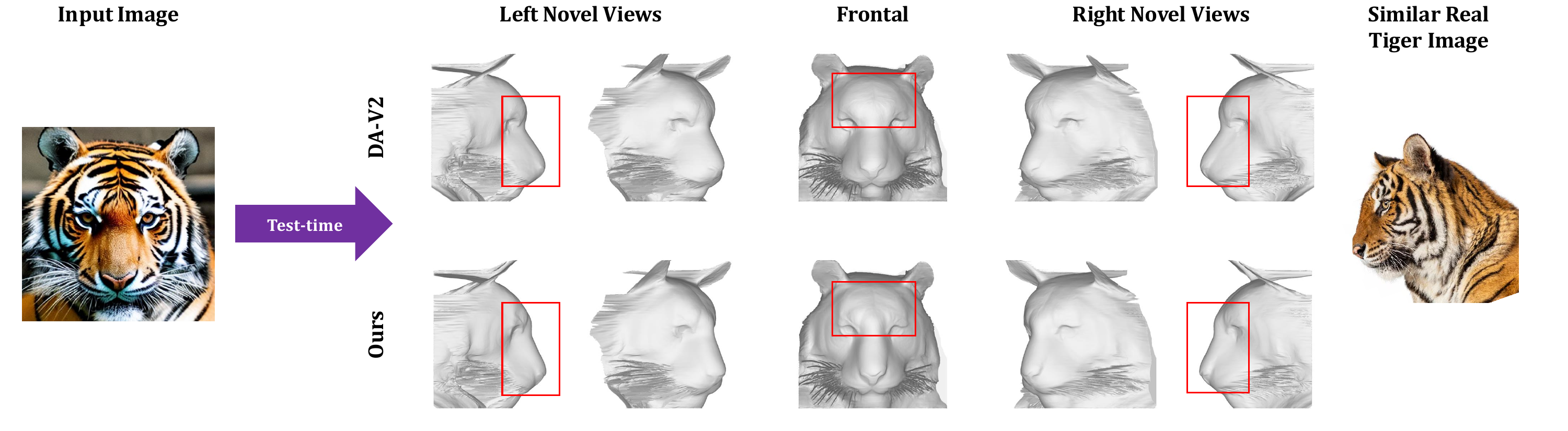}
\captionof{figure}{
    {\bf Re-Depth Anything} refines the predictions of Depth Anything V2~\cite{Yang2024DepthAV} by re-lighting the reconstructed geometry and extracting knowledge from diffusion models in a self-supervised manner. In this example, test-time optimization enhances facial detail (see frontal view) and refines the nose shape to look more like a tiger (side view), correcting the dog-like initial resemblance likely originating from a biased training distribution. The key contribution is a re-synthesis method that replaces photometric reconstruction for self-supervision.
}
\vspace{1em}
\label{fig:teaser}
}]

\begin{abstract}
\blfootnote{Code: \href{https://github.com/anantarb/Re-Depth-Anything}{https://github.com/anantarb/Re-Depth-Anything}}
Monocular depth estimation remains challenging, as foundation models such as Depth Anything V2 (DA-V2) struggle with real-world images that are far from the training distribution. We introduce Re-Depth Anything, a test-time self-supervision framework that bridges this domain gap by fusing foundation models with the powerful priors of large-scale 2D diffusion models. Our method performs label-free refinement directly on the input image by re-lighting the predicted depth map and augmenting the input. This re-synthesis method replaces classical photometric reconstruction by leveraging shape from shading (SfS) cues in a new, generative context with Score Distillation Sampling (SDS). To prevent optimization collapse, our framework updates only intermediate embeddings and the decoder’s weights, rather than optimizing the depth tensor directly or fine-tuning the full model. Across diverse benchmarks, Re-Depth Anything yields substantial gains in depth accuracy and realism over DA-V2, and applied on top of Depth Anything 3 (DA3) achieves state-of-the-art results, showcasing new avenues for self-supervision by geometric reasoning.
\end{abstract}
    
\section{Introduction}
\label{sec:intro}
Monocular Depth Estimation (MDE) aims to predict dense per-pixel depth from a single RGB image, enabling numerous applications, including 3D reconstruction~\cite{mildenhall2020nerf, Kerbl20233DGS, Ye2024GauStudioAM}, autonomous driving~\cite{Wang2018PseudoLiDARFV}, robotic navigation~\cite{Wofk2019FastDepthFM}, and virtual or augmented reality~\cite{Rasla2022TheRI}. 
Lately, high-quality depth maps for diverse images have been enabled by foundation models using Vision Transformers (ViTs)~\cite{Dosovitskiy2020AnII} with dense prediction heads~\cite{Ranftl2021VisionTF}.
Crucially, MiDaS~\cite{Ranftl2019TowardsRM} pioneered training on a myriad of labeled datasets, and
Depth Anything V2 (DA-V2)~\cite{Yang2024DepthAV} showed that sparse and often noisy depth labels can be enhanced with a teacher model trained on synthetic data. 
While these foundational models are setting new state-of-the-art performance,
inaccuracies in in-the-wild reconstruction remain (see Fig.~\ref{fig:teaser}).
Monocular depth estimation remains one of the fundamental yet challenging problems in computer vision. 

In this work, we introduce Re-Depth Anything, a test-time optimization framework designed to close the domain gap for feed forward models such as DA-V2 through self-supervision with 2D generative models. Given a single input image, our framework adapts the pre-trained model to the specific image content. The core idea is to re-light the feed-forward depth map under random illumination conditions and superimpose these renderings onto the input image. The depth map is then refined using a 2D diffusion model as a prior to score how realistic the augmented shading is. 
This plausibility estimate is backpropagated to the depth map through a differentiable renderer using the Blinn-Phong illumination model~\cite{blinn1977phong} and the SDS loss. Crucially, instead of directly optimizing the depth map or fine-tuning the entire network, we propose a targeted optimization strategy that jointly optimizes only the intermediate feature embeddings fed to the Dense Prediction Transformer (DPT) decoder and the decoder's weights. This approach preserves the strong geometric knowledge encoded in the embeddings while refining the final output.

3D knowledge distillation from 2D image diffusion models~\cite{Rombach2021HighResolutionIS, Podell2023SDXLIL, Saharia2022PhotorealisticTD} using the SDS loss and shading cues was pioneered by DreamFusion~\cite{Poole2022DreamFusionTU} for text-to-3D generation. %
Their key ingredient is optimizing a 3D NeRF representation~\cite{mildenhall2020nerf} such that its 2D renderings are perceived as realistic by the diffusion model. 
This and subsequent works~\cite{Wang2022ScoreJC, Lin2022Magic3DHT, Zhu2023HIFAHT, Lorraine2023ATT3DAT, Wang2023ProlificDreamerHA} have enabled reconstruction of real images by pairing virtual views with photometric reconstruction of the real one. However, this line of purely self-supervised learning from geometric relations suffers from cue ambiguities and lags behind supervised models. Our key advance is to apply the benefits of self-supervised learning on top of supervised methods by re-lighting the predicted depth map. This re-synthesis and augmentation of the input image is fundamentally different from photometric reconstruction with a full-fledged NeRF~\cite{mildenhall2020nerf} or Gaussian Splatting~\cite{Kerbl20233DGS} renderer in prior self-supervised work, and alleviates the problem of reconstructing appearance pixel-perfect.
Our main contributions are: \begin{itemize} 
\item Re-Depth Anything, a novel test-time optimization framework that adapts a pre-trained feed-forward model to real-world images using a 2D diffusion prior on re-synthesized depth predictions (w/o labeled data).
\item We propose a single-image re-lighting model that differentiably links the predicted depth map to the input image, enabling the use of an SDS loss for self-supervised geometry refinement from a single view.
\item We introduce a targeted optimization scheme that jointly optimizes the decoder’s input embeddings and its weights, which we show is crucial for avoiding overfitting and preserving geometric structure. 
\item The method is developed on DA-V2, and we verify generality on Depth Anything 3 (DA3)~\cite{depthanything3}.
 \end{itemize}

\section{Related Work}
\label{sec:relatedwork}

Our work, Re-Depth Anything, builds upon progress in three primary research areas: monocular depth estimation, test-time adaptation for monocular depth estimation, and the use of 2D diffusion models as priors for 3D reconstruction. Below, we review the most related methods from each of these domains.

\parag{Monocular Depth Estimation.} Monocular depth estimation has been a long-standing challenge in computer vision. Early approaches~\cite{Eigen2014DepthMP, Bhat2020AdaBinsDE, Li2022BinsFormerRA, Li2015DepthAS, Shao2023NDDepthNA, Yang2023GEDepthGE, Yuan2022NeuralWF, Xian2020StructureGuidedRL, Yin2019EnforcingGC} relied heavily on supervised learning, training on datasets with ground-truth depth, such as KITTI~\cite{Geiger2013VisionMR} for outdoor driving and NYU Depth V2~\cite{Silberman2012IndoorSA} for indoor scenes. 
More recently, the field has shifted towards building general-purpose foundation models for depth. MiDaS~\cite{Ranftl2019TowardsRM} enabled joint training on multiple datasets by predicting disparity instead of depth and by normalizing predictions to a unit range, 
enabling zero-shot generalization to new domains. Even better performance is possible with DPT heads~\cite{Ranftl2021VisionTF} and large diffusion models~\cite{ke2024repurposing, zhang2024betterdepth}.
DA~\cite{Yang2024DepthAU} and its successor, DA-V2~\cite{Yang2024DepthAV}, use the same relative prediction and achieved further improvements by training on massive-scale image datasets, aligning the predictions of a teacher model to sparse ground-truth measurements. This approach mitigates, but does not resolve, noise in LiDAR-based ground truth on in-the-wild images.

Another line of research predicts absolute depth without disparity normalization, from single images~\cite{yin2023metric3d, bochkovskii2024depth, wang2025moge, piccinelli2025unidepthv2} or multiple images~\cite{wang2024dust3r, leroy2024grounding, cabon2025must3r}. We focus on relative depth prediction, as these models typically yield higher surface detail, which we aim to improve. Moreover, the absolute depth scale is invariant to shading cues. Hence, we use the recent and popular DA-V2 model as our foundation, aiming to correct its errors rather than retraining it, so that it applies to underrepresented and out-of-distribution inputs.

\parag{Test-Time Adaptation for MDE.} %
Test-Time Adaptation (TTA) or Test-Time Optimization (TTO) aims to adapt a pre-trained model to a specific test input at inference time.

In the context of MDE, TTA often relies on self-supervision signals available from the input itself. For video inputs, temporal and photometric consistency between frames is a powerful signal used to fine-tune a depth network~\cite{Li2023OnSiteAF, Li2023TesttimeDA, Park2024TestTA}.

TTA for a single image is challenging because self-supervision cues are scarce. While existing methods rely on specific external priors---such as 3D meshes or sparse points---our method leverages a general-purpose 2D diffusion model to refine relative depth for any arbitrary scene.
Crucially, instead of relying solely on the input image's features, we introduce a powerful prior that provides a dense, geometry-aware supervisory signal for adaptation.

\parag{2D Diffusion Models as Priors for 3D.} 2D image diffusion models~\cite{Rombach2021HighResolutionIS, Podell2023SDXLIL, Saharia2022PhotorealisticTD}, trained on internet-scale image and text data, have learned incredibly rich priors about the visual world. A recent line of work has focused on leveraging these 2D priors for 3D tasks. The most influential is DreamFusion~\cite{Poole2022DreamFusionTU}, which introduced the SDS loss, enabling the use of a pre-trained text-to-image diffusion model as a loss function to optimize a 3D NeRF representation from scratch using only a text prompt. This concept has been extended and improved by numerous follow-ups~\cite{Wang2022ScoreJC, metzer2022latentnerf, Zhu2023HIFAHT, Lorraine2023ATT3DAT}, including using mesh representations~\cite{Lin2022Magic3DHT, Wang2023ProlificDreamerHA} and Gaussian Splatting~\cite{tang2023dreamgaussian, Chen2023Textto3DUG} as differentiable renderers.
Reconstructing a real input image brings additional challenges. RealFusion~\cite{MelasKyriazi2023RealFusion3R} proposes a method that combines real and synthetic views through diffusion-model fine-tuning and carefully selects compatible virtual views, while others directly fine-tune a 2D diffusion model for multi-view reconstruction~\cite{liu2023zero}.
Our re-lighting principle is inspired by the recent virtual shape-from-texture approach~\cite{bhattarai2025dreamtexture}, which utilizes shape-from-texture cues through virtual augmentation. However, these fully unsupervised methods have not demonstrated advantages over the latest supervised depth estimation techniques.

Our work builds directly on this idea but applies it in a novel context. Instead of generating a 3D shape or optimizing a NeRF from multiple views, we use the SDS loss on a single image to refine the parameters of a pre-trained, feed-forward depth estimation model (e.g., DA-V2) using re-lighting rather than photometric reconstruction. This test-time optimization adapts the model's prediction to the specific image content, guided by the diffusion model's knowledge about the shading of natural objects.

\parag{Single-View Geometry and Shading.} 
Shape-from-Shading (SfS)~\cite{horn1989shape, zhang1999shape} is a classical attempt to recover 3D shape from shading variations in a single 2D image. The idea is to decompose the image into (piecewise-constant) albedo and shading components, e.g.\ using the diffuse and specular components of a Blinn-Phong~\cite{blinn1977phong} model. However, classical SfS~\cite{horn1989shape, zhang1999shape} and other shape-from-X methods, such as Shape-from-Texture (SfT)~\cite{witkin1981recovering, white2006combining}, are highly ill-posed, relying on strong assumptions about lighting, texture regularity, and material properties that are rarely met in practice.

DreamFusion and RealFusion were the first to revisit the SfS principle in a modern context using generative models. Although they could lift some assumptions, RealFusion still follows the classical reconstruction principle of rendering a synthetic object optimized to reconstruct the input image. However, real illumination and material properties are complex, leading to artifacts around specular highlights when attempting to match them with simplistic shading models in deployed differentiable renderers.

By contrast, we do not attempt to solve the full, ill-posed inverse graphics problem. Instead, one of our key contributions is to augment the input with additional shading cues, which is achieved by modifying a simple, lightweight Blinn-Phong~\cite{blinn1977phong} renderer. 
This allows the diffusion prior to critique the plausibility of the 3D shape as expressed through its shading, linking the underlying geometry and image content without photometric reconstruction.

\section{Preliminaries and Notation}
\label{sec:preliminaries}

\paragraph{Score Distillation Sampling.}
Given a 2D image $\mathbf{X}$, rendered from a differentiable representation with parameters $\theta$, SDS~\cite{Poole2022DreamFusionTU} utilizes a pre-trained diffusion model $\phi$ to optimize $\theta$ via gradient descent. Specifically, a gradient toward a more likely image is obtained from the noise $\epsilon_{\phi}$ predicted by $\phi$ given a noisy image $\mathbf{X}_t$, text embedding $c$, and noise level $t$,
\begin{equation}
\label{eq:sds}
\nabla_{\theta}\mathcal{L}_{\text{SDS}}(\mathbf{X}, c) = \mathop{\mathbb{E}}_{\epsilon, t} \left[ w(t) \left( \epsilon_{\phi}(\mathbf{X}_t; c, t) - \epsilon \right) \frac{\partial \mathbf{X}}{\partial \theta} \right],
\end{equation}
where $w(t)$ is a weighting function and $\epsilon \sim \mathcal{N}(\mathbf{0}, \mathbf{I})$.
In DreamFusion~\cite{Poole2022DreamFusionTU}, the SDS loss matches 2D renderings from random angles to the text prompt. In our work, we apply the SDS loss to the re-lit image and use it to optimize both the intermediate feature embeddings and decoder weights in DA-V2 to enhance depth prediction.

\parag{Depth Anything V2 Architecture.}
DA-V2~\cite{Yang2024DepthAV} follows the recent trend of using ViT encoders, as in DINOv2~\cite{Oquab2023DINOv2LR} and DPT~\cite{Ranftl2021VisionTF}, with a DPT-based disparity decoder. Specifically, the encoder transforms input tokens into new feature representations using $L$ transformer layers. Features from four selected layers are extracted and passed to the DPT head for disparity prediction, with layer selection depending on the ViT variant. For example, layers $l={3, 6, 9, 12}$ are used in the ViT-Small configuration. Given an input image $\mathbf{I}$, we denote the extracted feature representations as embeddings $\mathbf{W}$. The final disparity $\mathbf{\hat{D}}_{\text{disp}} = f(\mathbf{W}; \theta)$ is predicted by the DPT head $f(\cdot)$, parameterized by pre-trained weights $\theta$ and taking embeddings $\mathbf{W}$ as input.

\begin{figure*}[t!]
    \centering
    \includegraphics[trim={0 0 0 0},width=\textwidth]{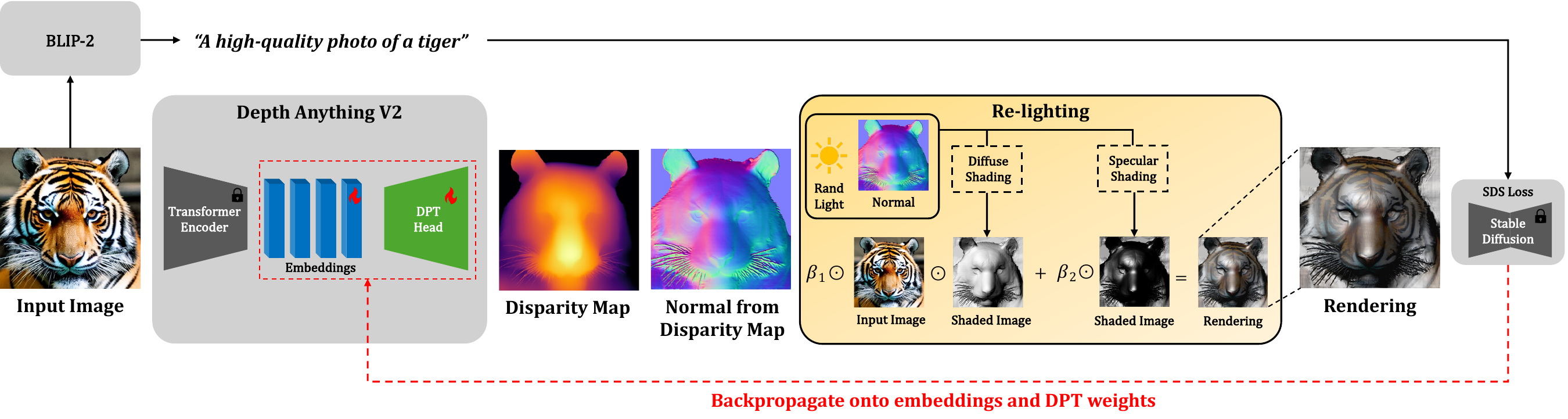}
    \caption{\textbf{Re-Depth Anything overview.} Our main contribution is the re-lighting module, which randomizes light conditions and shades the estimated geometry on the input. Notably, the re-lighting does not need to look physically accurate, as we are only augmenting the image, not photometrically reconstructing it. Key is also the optimization of the embeddings and decoder, while leaving the encoder frozen.}
    \label{fig:overview}
\end{figure*}

\section{Method}
\label{sec:method}

Given an input image $\mathbf{I} \in \mathbb{R}^{C \times H \times W}$, our goal is to refine an initial disparity map estimate $\mathbf{\hat{D}_{\text{init}}} \in \mathbb{R}^{H \times W}$ predicted by the pre-trained DA-V2~\cite{Yang2024DepthAV} model. As illustrated in Fig.~\ref{fig:overview}, Re-Depth Anything is a self-supervised test-time optimization framework that adapts the DA-V2 model to the specific input image.

Key to our method is how we use the SDS~\cite{Poole2022DreamFusionTU} loss as a 2D diffusion prior for 3D refinement. To this end, we first introduce a differentiable rendering function that links the predicted disparity map $\mathbf{\hat{D}}_\text{disp}$ to a re-illuminated image $\mathbf{\hat{I}}$ through augmentation. We then use the SDS loss on $\mathbf{\hat{I}}$ to jointly optimize the decoder’s input embeddings $\mathbf{W}$ and the weights $\theta$ to better align the re-illuminated and original images. We describe each component in detail below.

\subsection{Shaded Depth Rendering}
To leverage a 2D diffusion prior, we must establish a differentiable link between the depth map $\mathbf{\hat{D}}$ and a 2D image. A full inverse-rendering approach, which would decompose the scene into albedo, lighting, materials, and 3D geometry, is highly ill-posed from a single image.

Instead, we propose augmenting the input image with additional shading effects through re-lighting. We synthesize diffuse and specular reflectance maps with the classical Blinn-Phong shading model~\cite{blinn1977phong} as a function of the normals $\mathbf{N}$ of the depth map. Computing the normals at pixel coordinates $u,v$ requires a camera model. We test scaled orthographic and perspective projections with intrinsic matrices $\mK_\text{orth}$ and $\mK_\text{persp}$, respectively.
We then unproject the depth map element-wise into a 3D mesh with vertices
\begin{equation}
\mX = \mK_\text{persp}^{-1}
\begin{pmatrix}
\mU \mathbf{\hat{D}}\\
\mV \mathbf{\hat{D}}\\
\mathbf{\hat{D}}\\
\end{pmatrix}
\text{ or }
\mX = \mK_\text{orth}^{-1}
\begin{pmatrix}
\mU\\
\mV \\
\mathbf{\hat{D}}\\
\end{pmatrix}
,
\label{eq:unprojection}
\end{equation}
where $\mU$ and $\mV$ are the tensors of horizontal and vertical pixel coordinates ranging from $-1$ to $1$.
The normal $\mN$ is orthogonal to the spatial gradients $(\nabla \mX_u, \nabla \mX_v)$, which are computed across the entire image using the element-wise cross product.
\begin{equation}
\mN = \frac{\nabla \mX_v \times \nabla \mX_u}{||\nabla \mX_v \times \nabla \mX_u||_2}.
\end{equation}

Crucial for our re-lighting of the input image is using the inverse-tonemapped input image, $\tau^{-1}(\mathbf{I})$, as a proxy for the scene’s diffuse albedo, and we assume that specular highlights are colorless and not affected by albedo. %
While not physically accurate, this approach exploits the fact that illumination effects are linear, and there remains ambiguity between light and surface color, providing a sufficient, differentiable connection between geometry and re-lit appearance.
Specifically, we synthesize a re-illuminated image $\mathbf{\hat{I}} \in \mathbb{R}^{C \times H \times W}$ using Blinn-Phong shading~\cite{blinn1977phong},
\begin{equation}
\label{eq:rendering}
\mathbf{\hat{I}} = \tau\big(\beta_1 \max(\mathbf{N} \cdot \mathbf{l}, 0) \odot \tau^{-1}(\mathbf{I}) + \beta_2 \max(\mathbf{N} \cdot \mathbf{h}, 0)^{\alpha}\big),
\end{equation}
where $\mathbf{N} \in \mathbb{R}^{3 \times H \times W}$ is the per-pixel normal map derived from the depth gradients of $\mathbf{\hat{D}}$, $\mathbf{l} \in \mathbb{R}^3$ is the light direction, $\mathbf{h} \in \mathbb{R}^3$ is the halfway vector between $\mathbf{l}$ and the view direction $\mathbf{v} = [0, 0, 1]^T$, and $\alpha$, $\beta_1$, and $\beta_2$ are material and light-intensity parameters that are determined in the next section. The tone-mapping function $\tau(\mathbf{I}) = \mathbf{I}^{1/\gamma}$, with $\gamma = 2.2$, ensures that shading is performed in the linear RGB color space.

Note that using $\mathbf{I}$ as the albedo can double the shading effects (e.g.\ existing shadows get darker), and the addition of specular maps may saturate the image; however, this rarely occurs in our experiments and is similar to specular highlights appearing white in photographs. We keep values within bounds by clamping the rendered output $\mathbf{\hat{I}}$ to the range $[10^{-3}, 1]$.

\parag{Handling normalized relative depth.}
DA-V2 outputs normalized relative depths in the form of the normalized disparity map $\mathbf{\hat{D}}_\text{disp} = (1 / \mathbf{\hat{D}} - m)s$, where $m$ is the minimum disparity and $s$ is the inverse of the maximum--minimum disparity range. Converting to absolute depth requires
\begin{equation}
\label{eq:disparitytodepth}
    \mathbf{\hat{D}} = \frac{1}{\mathbf{\hat{D}}_\text{disp}/s + m} 
    = \frac{s}{\mathbf{\hat{D}}_\text{disp} + m s}, 
\end{equation}
where neither the scaling $s$ nor the offset $m$ is known at test time. However, the normal is invariant to global scale, and hence we only have to optimize the unknown scalar $b = ms$ alongside depth refinement.
Notably, we found the optimization to be insensitive to these parameters, likely because shading depends on the relative angle between the light and the normal, not absolute orientation. Specifically, it is sufficient to fix $b=0.1$ with scaled orthographic projection.

\subsection{Augmentation Objective}

Our goal is to refine the depth map to produce shading effects that yield plausible re-lightings of the input image. This augmentation principle lets us choose random light and material properties instead of estimating parameters for potentially absent or complex shading effects in the input image, as required for photometric reconstruction. At each optimization step, we randomly sample the light direction $\mathbf{l}$, diffuse and specular intensities $(\beta_1, \beta_2)$, and exponent $\alpha$ to ensure the refined geometry is consistent with the image across diverse shading conditions.

\parag{Loss Function.}
Plausibility of the augmentation is measured by the total loss, which combines the generative prior with a smoothness regularizer,%
\begin{equation}
\label{eq:loss}
\mathcal{L}(\mathbf{\hat{I}}, c, \mathbf{\hat{D}}_\text{disp}) = \mathcal{L}_{\text{SDS}}(\mathbf{\hat{I}}, c) + \frac{\lambda_1}{hw} \sum_{i,j} \| \Delta \mathbf{\hat{D}}^{i,j}_{\text{disp}} \|^2,
\end{equation}
where $\mathcal{L}_{\text{SDS}}$ is the SDS loss defined in Eq.~\eqref{eq:sds}, and $\mathbf{\hat{I}}$ is the image rendered from $\mathbf{\hat{D}}$ using Eq.~\eqref{eq:rendering}. The second term is an $\ell_1$ regularizer on the disparity gradients, which encourages smoother surfaces and prevents noisy artifacts.

To obtain the conditioning text prompt $c$, we employ BLIP-2~\cite{Li2023BLIP2BL}, a state-of-the-art image-to-text model, to generate a descriptive caption for the input image $\mathbf{I}$.

\subsection{Optimization Scheme}

Our goal is to refine the output depth map of feed-forward depth estimators, specifically the DA-V2 model. However, directly optimizing the depth map with the proposed re-lighting loss remains an ambiguous problem with many plausible solutions. Instead, we propose optimizing the latent feature space and weights of DA-V2, thereby leveraging the prior on 3D shapes learned at training time. %

Candidates for optimization are the entire DA-V2 weights or its components. We found that fine-tuning the entire DA-V2 tends to fall into poor local minima or causes the geometry to overfit to image textures.
To address this, we jointly optimize only the intermediate embeddings $\mathbf{W}$ (the intermediate feature embeddings of the frozen ViT encoder) and the DPT head’s weights $\theta$.
\begin{equation}
    \mathbf{W}^*, \theta^* = \arg\min_{\mathbf{W}, \theta} \mathcal{L}(\mathbf{\hat{I}}, c, \mathbf{\hat{D}}_\text{disp})
\end{equation}
where $\mathbf{\hat{D}}_\text{disp} = f(\mathbf{W}; \theta)$ and $\mathbf{\hat{I}}$ is a function of $\mathbf{\hat{D}}_\text{disp}$.

\parag{Depth Map Ensembling.}
The stochastic nature of the SDS loss, primarily due to random sampling of noise $\epsilon$ and timestep $t$, can lead to high variance in the optimization results. Consequently, disparity predictions can vary noticeably across different runs.
To stabilize the final prediction, inspired by the ensembling in Marigold~\cite{ke2024repurposing}, we perform the optimization $N$ times with different random seeds. We then aggregate the resulting disparity maps using a simple mean operation. The final disparity map $\mD_{\text{disp}}$ is obtained as
\begin{equation}
\mD_{\text{disp}} = \frac{1}{N}\sum_{i=1}^{N} f(\mathbf{W}_i^*; \theta_i^*),
\end{equation}
where $(\mathbf{W}_i^*, \theta_i^*)$ are the optimized embeddings and decoder weights from the $i$-th run.

\begin{table*}[t]
\caption{Comparison with DA-V2 across datasets. Relative error reduction of \textbf{Ours} over DA-V2 is shown in the last row of each dataset.}
\centering
\renewcommand{\arraystretch}{1.2}
\setlength{\tabcolsep}{5pt}
\resizebox{0.9\textwidth}{!}{
\begin{tabular}{l|lccc ccccccc} 
\hline
\multirow{2}{*}{Dataset} & \multirow{2}{*}{Method} 
 & \multicolumn{3}{c}{\textit{Higher is better} $\uparrow$} 
 &  & \multicolumn{6}{c}{\textit{Lower is better} $\downarrow$} \\ 
\cline{3-5} \cline{7-12}
 &  & $\delta_1$ & $\delta_2$ & $\delta_3$ 
 &  & AbsRel & RMSE & log10 & RMSE log & SI log & SqRel \\ 
\hline
\multirow{3}{*}{CO3D} 
 & DA-V2 & \textbf{1.0} & \textbf{1.0} & \textbf{1.0} &  & 0.00227 & 0.0602 & 0.000985 & 0.00321 & 0.321 & 0.000244 \\
 & \textbf{Ours + DA-V2} & \textbf{1.0} & \textbf{1.0} & \textbf{1.0} &  & \textbf{0.00223} & \textbf{0.0588} & \textbf{0.000968} & \textbf{0.00314} & \textbf{0.314} & \textbf{0.000235} \\
 \cline{2-12}
  & \textit{Rel. $\Delta$ (\%)} & - & - & - &  & \textbf{1.75} & \textbf{2.26} & \textbf{1.74} & \textbf{2.24} & \textbf{2.24} & \textbf{3.66} \\
\hline
\multirow{3}{*}{KITTI} 
 & DA-V2 & 0.568 & 0.796 & 0.902 &  & 0.305 & 7.01 & 0.118 & 0.348 & 33.6 & 2.49 \\
 & \textbf{Ours + DA-V2} & \textbf{0.593} & \textbf{0.818} & \textbf{0.917} &  & \textbf{0.283} & \textbf{6.71} & \textbf{0.110} & \textbf{0.319} & \textbf{30.7} & \textbf{2.20} \\
 \cline{2-12}
 & \textit{Rel. $\Delta$ (\%)} & \textbf{5.73} & \textbf{10.9} & \textbf{15.3} &  & \textbf{7.10} & \textbf{4.29} & \textbf{6.55} & \textbf{8.51} & \textbf{8.51} & \textbf{11.4} \\
\hline
\multirow{3}{*}{ETH3D} 
 & DA-V2 & 0.884 & 0.956 & 0.978 &  & 0.113 & 0.955 & 0.0448 & 0.153 & 15.1 & 0.391 \\
 & \textbf{Ours + DA-V2} & \textbf{0.898} & \textbf{0.965} & \textbf{0.982} &  & \textbf{0.104} & \textbf{0.875} & \textbf{0.0413} & \textbf{0.143} & \textbf{14.1} & \textbf{0.347} \\
 \cline{2-12}
 & \textit{Rel. $\Delta$ (\%)} & \textbf{12.2} & \textbf{21.1} & \textbf{19.5} &  & \textbf{8.30} & \textbf{8.39} & \textbf{7.72} & \textbf{6.44} & \textbf{6.22} & \textbf{11.1} \\
\hline
\end{tabular}
}
\label{tab:quantitative}
\end{table*}

\section{Experiments}
\label{sec:intro}

We study the effectiveness and accuracy of our self-supervised re-lighting approach on three benchmarks, demonstrating consistent improvements over the DA-V2 and DA3 baselines across established metrics, including a relative error improvement of up to 11.4\% and a normal map improvement of $14.7\%$. Fig.~\ref{fig:qualitative} shows exemplary cases that are improved by removing noise from flat areas and adding missing details. Additional results are shown in the supplemental document.

\parag{Baselines.} We utilize the small variant of the DA-V2 architecture as our base model, which is one of the most popular monocular relative depth estimation methods. Besides various baseline variants that use only parts of our contributions, we also compare against a simple Shape-from-Shading implementation to demonstrate the advantage of re-lighting over photometric reconstruction.

\parag{Implementation Details.}
For all experiments, input images are first resized, maintaining their original aspect ratio, such that at least one side measures 518 pixels to match the training resolution of the DA-V2 model. Before applying Stable Diffusion, the images are further zero-padded if they have a non-square aspect ratio and resized to 512 $\times$ 512.

Our entire pipeline is implemented in PyTorch. For SDS guidance, we employ v1.5 of Stable Diffusion~\cite{Rombach2021HighResolutionIS}. We optimize the encoder's embeddings and DPT weights for 1000 iterations using the AdamW optimizer. We set a learning rate of $1 \times 10^{-3}$ for the embeddings and $2 \times 10^{-6}$ for the DPT weights. We set the regularization weight $\lambda_1$ to $1.0$. At each optimization step, we uniformly sample two coefficients $(\beta_1, \beta_2) \sim \mathcal{U}[0, 1]$ and an exponent $\alpha$, where $\alpha=2^k$ and $k \sim \mathcal{U}[2, 8]$. $\beta_1$ and $\beta_2$ are subsequently normalized to ensure their sum is $1.0$ (i.e.\ $\beta_1 + \beta_2 = 1$). Similarly, the light direction vector $\mathbf{l} = (L_x, L_y, L_z)$ is sampled by drawing the X and Y coordinates from a uniform distribution, $L_x, L_y \sim \mathcal{U}[-1, 1]$, while fixing the Z coordinate to $L_z=1$. The resulting vector $\mathbf{l}$ is then $\ell_2$-normalized. In the absence of known camera parameters for in-the-wild images, our default is a scaled orthographic camera, and we optimize the scaling starting from 7.0. For a perspective camera, the focal length is initialized to 2.0 and refined alongside disparity optimization.
The final prediction is generated by aggregating the results from 10 optimization runs. Each of these 10 runs is initialized from the original pre-trained weights of the DA-V2 model. We conduct our evaluation at the resolution of the initial 518-pixel-side resized input image. One run takes approximately 80 seconds on a single NVIDIA RTX 5000.

\parag{Datasets.}
We evaluate Re-Depth Anything on three standard benchmarking datasets. CO3Dv2~\cite{reizenstein21co3d} contains multi-view images of several objects across 50 categories, with camera poses, 3D point clouds, foreground masks, and depth maps that we utilize as sparse ground truth depth. From each sequence with a valid depth map, we randomly selected two images. This preprocessing step yielded a total of 80 images from 20 object categories.
KITTI~\cite{Geiger2013VisionMR} is a large-scale autonomous driving dataset featuring sparse metric depth captured by a LiDAR sensor. We randomly sampled 10 images from each sequence in the official validation set, resulting in a total of 130 images.
ETH3D~\cite{schops2019bad} is a high-resolution benchmark with both indoor and outdoor scenes. We randomly sampled ten raw images and their corresponding depth maps from 13 scenes. This process resulted in a total of 130 images for evaluation.

\parag{Evaluation protocol.}
We apply the widely adopted metrics for assessing the quality of monocular depth estimation: $\delta_1$, $\delta_2$, $\delta_3$, AbsRel, RMSE, log10, RMSE log, SI log, and SqRel. We compute these metrics on absolute depth maps, obtained by first finding the least-squares affine fit to the labels in disparity, then converting to depth, and subsequently finding the affine fit in depth, as is typical for relative depth prediction on these benchmarks~\cite{Yang2024DepthAV}. 

\subsection{Quantitative Evaluation}

Table~\ref{tab:quantitative} shows that our test-time refinement improves upon DA-V2 across CO3D, KITTI, and ETH3D on all nine evaluation metrics. Notably, we achieve significant relative reductions in error, including 8.5\% in SI log and RMSE log on KITTI, alongside an 8.4\% reduction in AbsRel on ETH3D.

On CO3D, errors are smaller overall, leading to smaller yet still consistent improvements. This robustness across nine different metrics validates the efficacy of our self-supervised approach, demonstrating considerable potential for fine-tuning foundational models.

Notably, the improvements are consistent across CO3D, which covers single objects in a close-up setting, street scenes in the KITTI dataset, and indoor scenes in the ETH3D dataset. This highlights the strong generalization capability of our test-time re-lighting approach, inherited from the robustness of generative image models. The nature of the improvement is best explained with examples.\\

\begin{table}[h]
    \caption{\textbf{Quantitative comparison against DA3.}}
    \centering
    \renewcommand{\arraystretch}{1.2}
    \setlength{\tabcolsep}{5pt}
    \resizebox{0.95\linewidth}{!}{
    \begin{tabular}{l|lccc} 
    \hline
    Dataset & Method & AbsRel  & SqRel & Normal MSE \\ 
    \hline
    \multirow{3}{*}{CO3D} & DA3 & 0.00251 & 0.000317 & 0.000479  \\
    & \textbf{Ours + DA3} & \textbf{0.00238} & \textbf{0.000294} & \textbf{0.000409}  \\
    \cline{2-5}
    & \textit{Rel. $\Delta$ (\%)} & \textbf{4.83} & \textbf{7.39} & \textbf{14.65} \\
    \hline
    \multirow{3}{*}{ETH3D} & DA3 & 0.0444 & 0.0853 & 0.00057  \\
    & \textbf{Ours + DA3} & \textbf{0.0441} & \textbf{0.0847} & \textbf{0.00054}  \\
    \cline{2-5}
    & \textit{Rel. $\Delta$ (\%)} & \textbf{0.54} & \textbf{0.71} & \textbf{3.74} \\
    \hline
    \end{tabular}
    }
    \label{tab:ablation-dav3}
\end{table}

\noindent \textbf{Re-Depth generalization to DA3.} In Table~\ref{tab:ablation-dav3}, we show the effectiveness of our self-supervised refinement strategy on a different backbone by applying it to the recent DA3~\cite{depthanything3} model (DA3MONO-LARGE). Our method consistently improves DA3 on CO3D and ETH3D, achieving state-of-the-art results. In particular, normal errors are improved by up to $15\%$, verifying strong detail improvement. Further results and details are given in the supplemental document.

\subsection{Qualitative Evaluation}

\begin{figure*}[p]
    \centering
    \includegraphics[width=\textwidth]{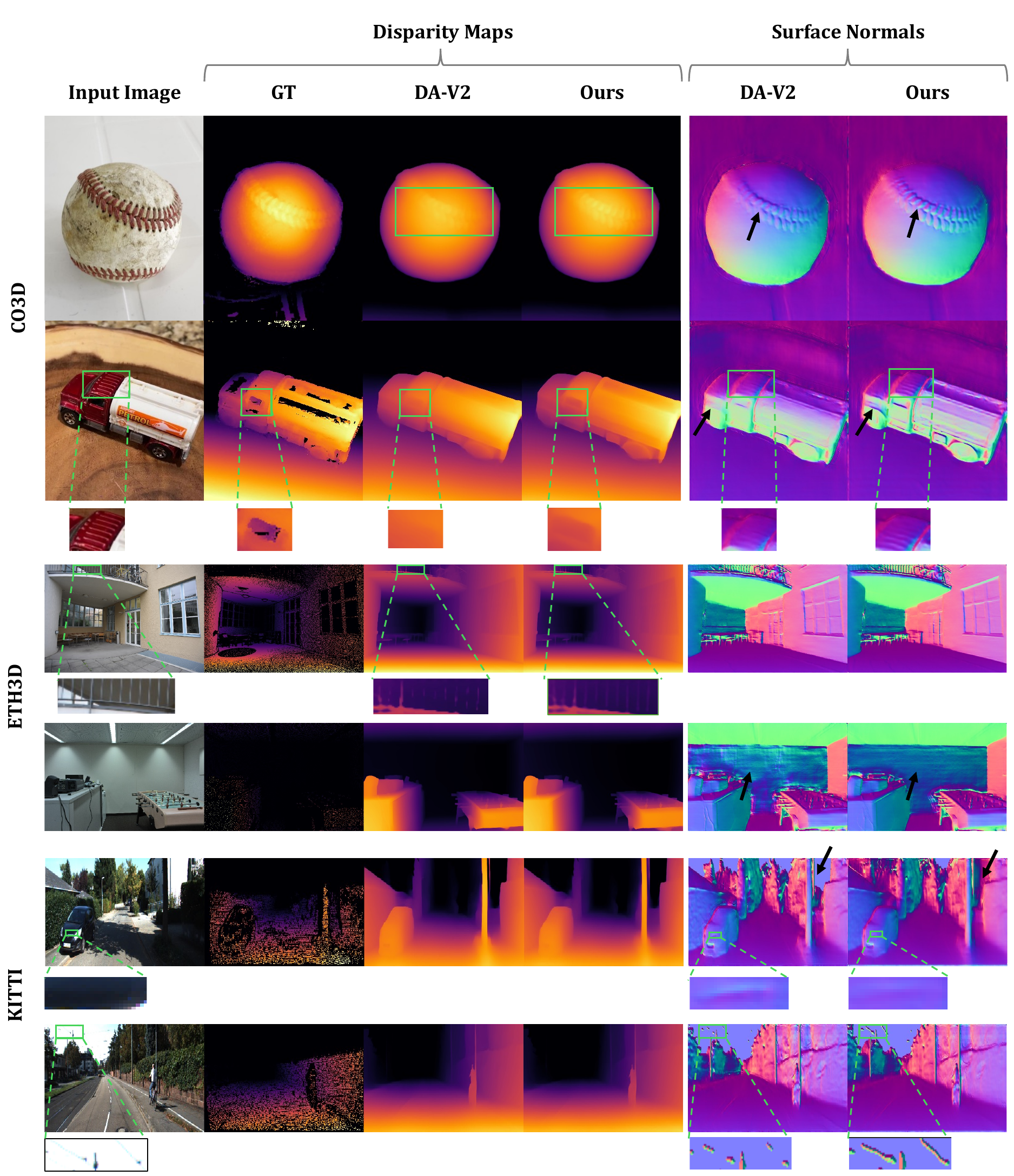}
    \caption{\textbf{Qualitative Comparison,} highlighting the added detail (rows 1,2,3,6) and noise-removal effects (rows 4,5).}
    \label{fig:qualitative}
\end{figure*}

For visual assessment, we present a qualitative comparison against DA-V2 in Fig.~\ref{fig:qualitative}. Re-Depth Anything produces visibly superior results by enhancing fine-grained details such as the threads on a ball (first image), balcony railings, and electricity wires (second-to-last image), while also removing noise from flat surfaces, as indicated by an arrow in the fourth example. These qualitative improvements are consistent with the quantitative gains reported in Table~\ref{tab:quantitative}.

\begin{figure}[h]
    \centering
    \includegraphics[width=0.9\linewidth, trim={0.6cm 0.4cm 0.6cm 0.4cm},clip]{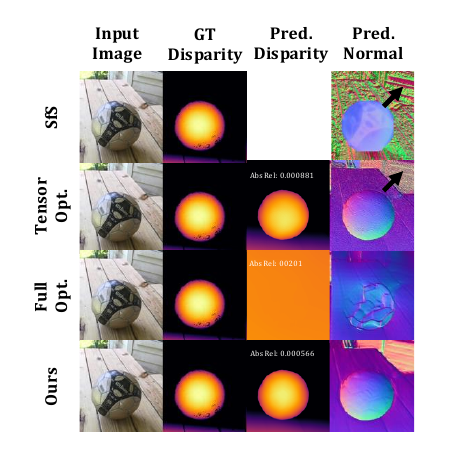}
    \caption{\textbf{Qualitative ablation} showing that directly optimizing depth or fine-tuning the whole network at once is detrimental. The listed error values relate to visual and quantitative improvements.}
    \label{fig:ablation}
\end{figure}

We also compare to classical shape-from-shading, which fails drastically when its strict assumptions are violated. For instance, even in the relatively simple ball example in Fig.~\ref{fig:ablation} (first row), discoloration of the leather leads to spurious and noisy normals. This highlights the importance of our re-lighting augmentation strategy, which does not assume albedo constancy and does not suffer from seam artifacts typically associated with shape-from-shading.

\FloatBarrier

\parag{Limitations.} We rarely observed small hallucinated edges, such as the sticker on the truck in Fig.~\ref{fig:qualitative}, which are plausible but incorrect. Sometimes, the method extends geometry into the sky and over-smooths fine details, such as trees in

\begin{wrapfigure}{r}{0.45\linewidth}
    \centering
    \includegraphics[width=\linewidth,
        trim={5cm 1.7cm 3cm 1cm},clip]{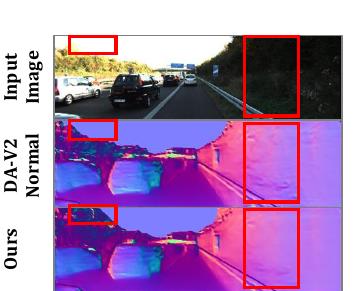}
    \caption{\textbf{Limitations.}}
    \label{fig:limitation}
\end{wrapfigure}%
\noindent
dark regions, as seen in the KITTI example (see inset Fig.~\ref{fig:limitation}), which could potentially be handled by thresholding. In general, we found that single objects in the CO3D dataset show stronger detail improvements, while in room and street scenes the largest gains come from removing suspicious details in the initial DA-V2 predictions that lead to unrealistic re-lighting highlights and are thus effectively removed by our method while preserving actual detail.

\subsection{Ablation study}

\paragraph{Optimization.} Table~\ref{tab:ablation} presents an ablation study on the CO3D evaluation set, comparing our full pipeline to a baseline lacking the SDS loss and four optimization variants: (1) direct depth pixel optimization, (2) full DA-V2 fine-tuning, (3) fine-tuning only embeddings and DPT weights, and (4) a two-stage version of (3) that first optimizes the embeddings and subsequently fine-tunes the decoder.

The qualitative results in Fig.~\ref{fig:ablation} show that direct optimization (1) creates noise artifacts (first row), while full fine-tuning (2) causes the geometry to collapse (second row), even though we reduced the learning rate to a very small value of $2 \times 10^{-6}$ for optimizing both the ViT encoder and the DPT decoder.
Our chosen approach (3) strikes a balance, enhancing detail without degradation. These visual observations are corroborated by the quantitative metrics in Table~\ref{tab:ablation}. Although variants (3) and (4) are visually comparable, (3) achieves lower errors.

\begin{figure}[t]
    \centering
    \includegraphics[width=0.95\linewidth]{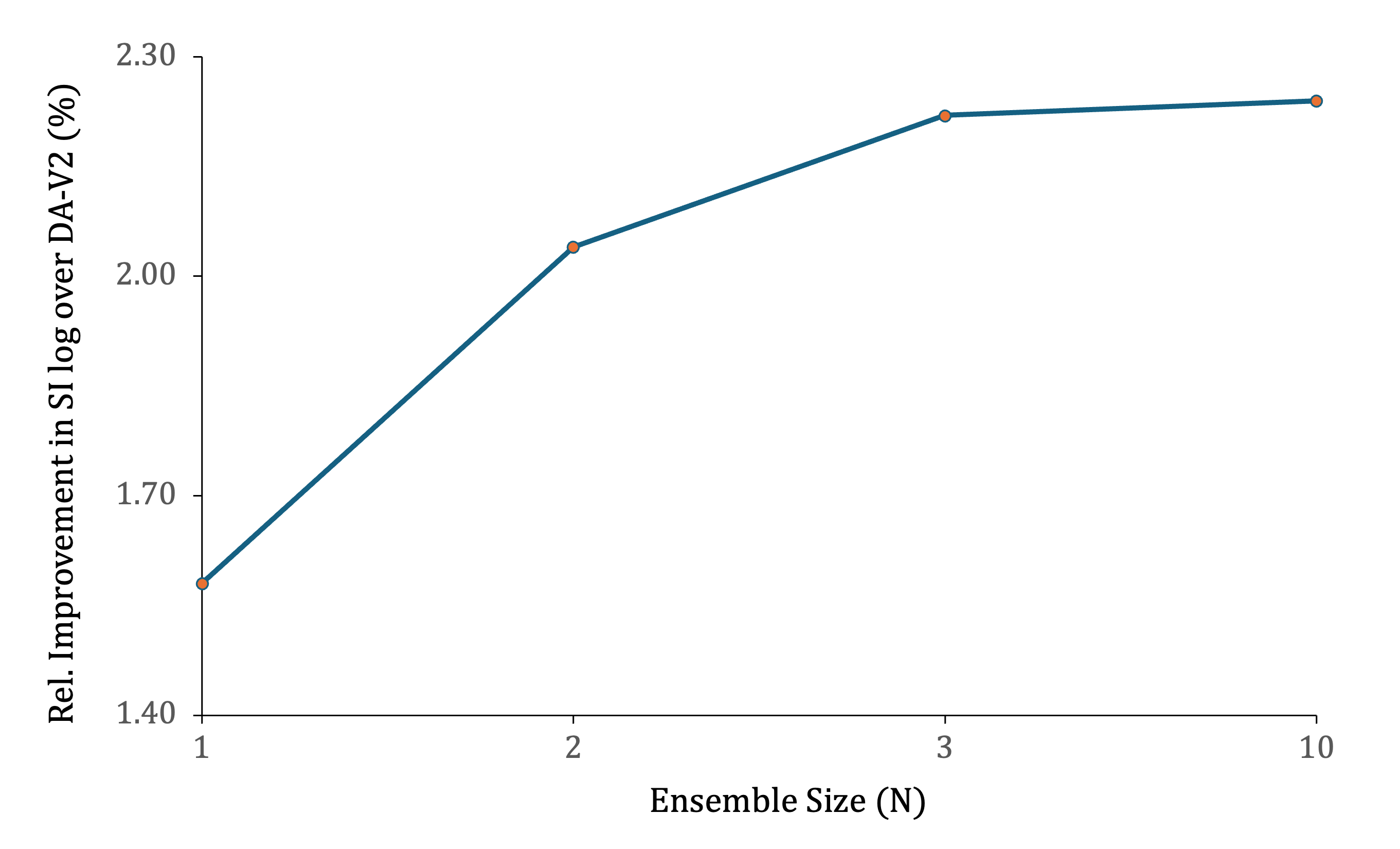}
    \caption{\textbf{Ensemble size} vs.~improvement in SI log on CO3D.}
    \label{fig:ensemble}
\end{figure}

\parag{Ensembling predictions.} We investigate the impact of ensembling predictions via mean aggregation. The results, shown in Fig.~\ref{fig:ensemble}, demonstrate clear benefits but with rapidly diminishing returns. While a prediction from a single run achieves a 1.58\% improvement in SI log over DA-V2, ensembling predictions from three runs significantly increases this to 2.22\%. This benefit quickly saturates, with 10 predictions offering only a negligible further gain (2.24\%).

\begin{table}[t]
    \caption{\textbf{Ablation on the CO3D dataset} showing the significance of the major design choices.}
    \centering
    \renewcommand{\arraystretch}{1.2}
    \setlength{\tabcolsep}{5pt}
    \resizebox{0.99\linewidth}{!}{
    \begin{tabular}{lccccccc} 
    \hline
    Method & AbsRel $\downarrow$ & RMSE $\downarrow$ & log10 $\downarrow$ & RMSE log $\downarrow$ & SI log $\downarrow$ & SqRel $\downarrow$ \\ 
    \hline
    w/o $\mathcal{L}_{\text{SDS}}$ & 0.00427 & 0.0993 & 0.00185\phantom{0} & 0.00532 & 0.532 & 0.000661 \\
    Tensor Model & 0.00226 & 0.0601 & 0.000985 & 0.00321 & 0.321 & 0.000244 \\
    Full Model & 0.00331 & 0.0779 & 0.00143\phantom{0} & 0.00418 & 0.418 & 0.000412\\
    Two Stage & 0.00225 & 0.0597 & 0.000979 & 0.00319 & 0.319 & 0.000241 \\
    \textbf{Ours} & \textbf{0.00223} & \textbf{0.0588} & \textbf{0.000968} & \textbf{0.00314} & \textbf{0.314} & \textbf{0.000235} \\
    \hline
    \end{tabular}
    }
    \label{tab:ablation}
\end{table}

\section{Conclusion}

Re-Depth Anything presents a new method for test-time optimization through re-lighting. The key contribution is using generative models to score shading--image alignment instead of requiring photometric reconstruction. This alleviates the need to construct a photorealistic renderer for inverse graphics and shows consistent improvements across all tested datasets and metrics. Notably, it sets a new state-of-the-art for detail reconstruction on the CO3D dataset using DA3 as the backbone.

\section*{Acknowledgment}
We thank all members of the Visual AI for Extended Reality lab at Bielefeld University for providing crucial feedback and participating in helpful discussions.

{
    \small
    \bibliographystyle{ieeenat_fullname}
    \bibliography{main}

@String(CVPR= {IEEE Conf. Comput. Vis. Pattern Recog.})

@String(ICCV= {Int. Conf. Comput. Vis.})

@String(ECCV= {Eur. Conf. Comput. Vis.})

@String(BMVC= {Brit. Mach. Vis. Conf.})

@String(TOG= {ACM Trans. Graph.})

@String(ICLR = {Int. Conf. Learn. Represent.})

@String(CVPR  = {CVPR})

@String(ICCV  = {ICCV})

@String(ECCV  = {ECCV})

@String(BMVC  =	{BMVC})

@String(TOG   = {ACM TOG})

@String(ICLR  = {ICLR})

@article{witkin1981recovering,
  title={Recovering surface shape and orientation from texture},
  author={Witkin, Andrew P.},
  journal={Artificial Intelligence},
  volume={17},
  number={1-3},
  pages={17--45},
  year={1981},
  publisher={Elsevier},
  doi={10.1016/0004-3702(81)90026-6}
}

@article{bhattarai2025dreamtexture,
  title={DreamTexture: Shape from Virtual Texture with Analysis by Augmentation},
  author={Bhattarai, Ananta R and He, Xingzhe and Sheffer, Alla and Rhodin, Helge},
  journal={ArXiv},
  year={2025}
}

@inproceedings{zhang2024betterdepth,
  title={Betterdepth: Plug-and-play diffusion refiner for zero-shot monocular depth estimation},
  author={Zhang, Xiang and Ke, Bingxin and Riemenschneider, Hayko and Metzger, Nando and Obukhov, Anton and Gross, Markus and Schindler, Konrad and Schroers, Christopher},
  booktitle={Neural Information Processing Systems (NeurIPS)},
  year={2024}
}

@inproceedings{wang2024dust3r,
  title={Dust3r: Geometric 3d vision made easy},
  author={Wang, Shuzhe and Leroy, Vincent and Cabon, Yohann and Chidlovskii, Boris and Revaud, Jerome},
  booktitle={Conference on Computer Vision and Pattern Recognition (CVPR)},
  year={2024}
}

@inproceedings{yin2023metric3d, 
    title={Metric3d: Towards zero-shot metric 3d prediction from a single image}, author={Yin, Wei and Zhang, Chi and Chen, Hao and Cai, Zhipeng and Yu, Gang and Wang, Kaixuan and Chen, Xiaozhi and Shen, Chunhua}, 
    booktitle={International Conference on Computer Vision (ICCV)},
    year={2023} 
}

@inproceedings{piccinelli2025unidepthv2,
  title={Unidepthv2: Universal monocular metric depth estimation made simpler},
  author={Piccinelli, Luigi and Sakaridis, Christos and Yang, Yung-Hsu and Segu, Mattia and Li, Siyuan and Abbeloos, Wim and Van Gool, Luc},
  booktitle={Conference on Computer Vision and Pattern Recognition (CVPR)},
  year={2025}
}

@inproceedings{cabon2025must3r,
  title={Must3r: Multi-view network for stereo 3d reconstruction},
  author={Cabon, Yohann and Stoffl, Lucas and Antsfeld, Leonid and Csurka, Gabriela and Chidlovskii, Boris and Revaud, Jerome and Leroy, Vincent},
  booktitle={Computer Vision and Pattern Recognition Conference (CVPR)},
  year={2025}
}

@inproceedings{leroy2024grounding,
  title={Grounding image matching in 3d with mast3r},
  author={Leroy, Vincent and Cabon, Yohann and Revaud, J{\'e}r{\^o}me},
  booktitle={European Conference on Computer Vision (ECCV)},
  year={2024},
}

@inproceedings{ke2024repurposing,
  title={Repurposing diffusion-based image generators for monocular depth estimation},
  author={Ke, Bingxin and Obukhov, Anton and Huang, Shengyu and Metzger, Nando and Daudt, Rodrigo Caye and Schindler, Konrad},
  booktitle={Computer Vision and Pattern Recognition Conference (CVPR)},
  year={2024}
}

@inproceedings{wang2025moge,
  title={Moge: Unlocking accurate monocular geometry estimation for open-domain images with optimal training supervision},
  author={Wang, Ruicheng and Xu, Sicheng and Dai, Cassie and Xiang, Jianfeng and Deng, Yu and Tong, Xin and Yang, Jiaolong},
  booktitle={Computer Vision and Pattern Recognition Conference (CVPR)},
  year={2025}
}

@inproceedings{bochkovskii2024depth,
  title={Depth pro: Sharp monocular metric depth in less than a second},
  author={Aleksei Bochkovskii and Amaël Delaunoy and Hugo Germain and Marcel Santos and Yichao Zhou and Stephan R. Richter and Vladlen Koltun},
  booktitle={International Conference on Learning Representations (ICLR)},
  year={2025}
}

@inproceedings{Eigen2014DepthMP,
  title={Depth Map Prediction from a Single Image using a Multi-Scale Deep Network},
  author={David Eigen and Christian Puhrsch and Rob Fergus},
  booktitle={Neural Information Processing Systems (NeurIPS)},
  year={2014},
  url={https://api.semanticscholar.org/CorpusID:2255738}
}

@article{Geiger2013VisionMR,
  title={Vision meets robotics: The KITTI dataset},
  author={Andreas Geiger and Philip Lenz and Christoph Stiller and Raquel Urtasun},
  journal={The International Journal of Robotics Research},
  year={2013},
  volume={32},
  pages={1231 - 1237},
  url={https://api.semanticscholar.org/CorpusID:9455111}
}

@inproceedings{Bhat2020AdaBinsDE,
  title={AdaBins: Depth Estimation Using Adaptive Bins},
  author={Shariq Farooq Bhat and Ibraheem Alhashim and Peter Wonka},
  booktitle={Conference on Computer Vision and Pattern Recognition (CVPR)},
  year={2020},
  url={https://api.semanticscholar.org/CorpusID:227227779}
}

@article{Li2022BinsFormerRA,
  title={BinsFormer: Revisiting Adaptive Bins for Monocular Depth Estimation},
  author={Zhenyu Li and Xuyang Wang and Xianming Liu and Junjun Jiang},
  journal={IEEE Transactions on Image Processing},
  year={2022},
  volume={33},
  pages={3964-3976},
  url={https://api.semanticscholar.org/CorpusID:247939833}
}

@inproceedings{Li2015DepthAS,
  title={Depth and surface normal estimation from monocular images using regression on deep features and hierarchical CRFs},
  author={Bo Li and Chunhua Shen and Yuchao Dai and Anton van den Hengel and Mingyi He},
  booktitle={Conference on Computer Vision and Pattern Recognition (CVPR)},
  year={2015},
  url={https://api.semanticscholar.org/CorpusID:206592782}
}

@inproceedings{Yang2024DepthAU,
  title={Depth Anything: Unleashing the Power of Large-Scale Unlabeled Data},
  author={Lihe Yang and Bingyi Kang and Zilong Huang and Xiaogang Xu and Jiashi Feng and Hengshuang Zhao},
  booktitle={Conference on Computer Vision and Pattern Recognition (CVPR)},
  year={2024},
  url={https://api.semanticscholar.org/CorpusID:267061016}
}

@inproceedings{Shao2023NDDepthNA,
  title={NDDepth: Normal-Distance Assisted Monocular Depth Estimation},
  author={Shuwei Shao and Zhongcai Pei and Weihai Chen and Xingming Wu and Zhengguo Li},
  booktitle={International Conference on Computer Vision (ICCV)},
  year={2023},
  url={https://api.semanticscholar.org/CorpusID:262055086}
}

@inproceedings{Yang2024DepthAV,
  title={Depth Anything V2},
  author={Lihe Yang and Bingyi Kang and Zilong Huang and Zhen Zhao and Xiaogang Xu and Jiashi Feng and Hengshuang Zhao},
  booktitle={Neural Information Processing Systems (NeurIPS)},
  year={2024},
  url={https://api.semanticscholar.org/CorpusID:270440448}
}

@inproceedings{Saharia2022PhotorealisticTD,
  title={Photorealistic Text-to-Image Diffusion Models with Deep Language Understanding},
  author={Chitwan Saharia and William Chan and Saurabh Saxena and Lala Li and Jay Whang and Emily L. Denton and Seyed Kamyar Seyed Ghasemipour and Burcu Karagol Ayan and Seyedeh Sara Mahdavi and Raphael Gontijo Lopes and Tim Salimans and Jonathan Ho and David J. Fleet and Mohammad Norouzi},
  booktitle={Neural Information Processing Systems (NeurIPS)},
  year={2022},
}

@inproceedings{Silberman2012IndoorSA,
  title={Indoor Segmentation and Support Inference from RGBD Images},
  author={Nathan Silberman and Derek Hoiem and Pushmeet Kohli and Rob Fergus},
  booktitle={European Conference on Computer Vision (ECCV)},
  year={2012},
  url={https://api.semanticscholar.org/CorpusID:545361}
}

@article{Oquab2023DINOv2LR,
  title={DINOv2: Learning Robust Visual Features without Supervision},
  author={Maxime Oquab and Timoth{\'e}e Darcet and Th{\'e}o Moutakanni and Huy Q. Vo and Marc Szafraniec and Vasil Khalidov and Pierre Fernandez and Daniel Haziza and Francisco Massa and Alaaeldin El-Nouby and Mahmoud Assran and Nicolas Ballas and Wojciech Galuba and Russ Howes and Po-Yao (Bernie) Huang and Shang-Wen Li and Ishan Misra and Michael G. Rabbat and Vasu Sharma and Gabriel Synnaeve and Huijiao Xu and Herv{\'e} J{\'e}gou and Julien Mairal and Patrick Labatut and Armand Joulin and Piotr Bojanowski},
  journal={ArXiv},
  year={2023},
  url={https://api.semanticscholar.org/CorpusID:258170077}
}

@inproceedings{Rombach2021HighResolutionIS,
  title={High-Resolution Image Synthesis with Latent Diffusion Models},
  author={Robin Rombach and A. Blattmann and Dominik Lorenz and Patrick Esser and Bj{\"o}rn Ommer},
  booktitle={Conference on Computer Vision and Pattern Recognition (CVPR)},
  year={2021},
  url={https://api.semanticscholar.org/CorpusID:245335280}
}

@inproceedings{Poole2022DreamFusionTU,
  title={DreamFusion: Text-to-3D using 2D Diffusion},
  author={Ben Poole and Ajay Jain and Jonathan T. Barron and Ben Mildenhall},
  booktitle={International Conference on Learning Representations (ICLR)},
  year={2023},
  url={https://api.semanticscholar.org/CorpusID:252596091}
}

@inproceedings{Ranftl2021VisionTF,
  title={Vision Transformers for Dense Prediction},
  author={Ren{\'e} Ranftl and Alexey Bochkovskiy and Vladlen Koltun},
  booktitle={International Conference on Computer Vision (ICCV)},
  year={2021},
  url={https://api.semanticscholar.org/CorpusID:232352612}
}

@inproceedings{Li2023BLIP2BL,
  title={BLIP-2: Bootstrapping Language-Image Pre-training with Frozen Image Encoders and Large Language Models},
  author={Junnan Li and Dongxu Li and Silvio Savarese and Steven C. H. Hoi},
  booktitle={International Conference on Machine Learning (ICML)},
  year={2023},
  url={https://api.semanticscholar.org/CorpusID:256390509}
}

@inproceedings{Wang2022ScoreJC,
  title={Score Jacobian Chaining: Lifting Pretrained 2D Diffusion Models for 3D Generation},
  author={Haochen Wang and Xiaodan Du and Jiahao Li and Raymond A. Yeh and Gregory Shakhnarovich},
  booktitle={Conference on Computer Vision and Pattern Recognition (CVPR)},
  year={2022},
  url={https://api.semanticscholar.org/CorpusID:254125253}
}

@inproceedings{Yang2023GEDepthGE,
  title={GEDepth: Ground Embedding for Monocular Depth Estimation},
  author={Xiaodong Yang and Zhuang Ma and Zhiyu Ji and Zhe Ren},
  booktitle={International Conference on Computer Vision (ICCV)},
  year={2023},
  url={https://api.semanticscholar.org/CorpusID:262044589}
}

@inproceedings{Lin2022Magic3DHT,
  title={Magic3D: High-Resolution Text-to-3D Content Creation},
  author={Chen-Hsuan Lin and Jun Gao and Luming Tang and Towaki Takikawa and Xiaohui Zeng and Xun Huang and Karsten Kreis and Sanja Fidler and Ming-Yu Liu and Tsung-Yi Lin},
  booktitle={Conference on Computer Vision and Pattern Recognition (CVPR)},
  year={2022},
  url={https://api.semanticscholar.org/CorpusID:253708074}
}

@inproceedings{Podell2023SDXLIL,
  title={SDXL: Improving Latent Diffusion Models for High-Resolution Image Synthesis},
  author={Dustin Podell and Zion English and Kyle Lacey and A. Blattmann and Tim Dockhorn and Jonas Muller and Joe Penna and Robin Rombach},
  booktitle={International Conference on Learning Representations (ICLR)},
  year={2023},
  url={https://api.semanticscholar.org/CorpusID:259341735}
}

@inproceedings{Wofk2019FastDepthFM,
  title={FastDepth: Fast Monocular Depth Estimation on Embedded Systems},
  author={Diana Wofk and Fangchang Ma and Tien-Ju Yang and Sertac Karaman and Vivienne Sze},
  booktitle={International Conference on Robotics and Automation (ICRA)},
  year={2019},
  url={https://api.semanticscholar.org/CorpusID:72940839}
}

@inproceedings{Zhu2023HIFAHT,
  title={HIFA: High-fidelity Text-to-3D Generation with Advanced Diffusion Guidance},
  author={Junzhe Zhu and Peiye Zhuang and Oluwasanmi Koyejo},
  booktitle={International Conference on Learning Representations (ICLR)},
  year={2023},
  url={https://api.semanticscholar.org/CorpusID:258967476}
}

@inproceedings{Yuan2022NeuralWF,
  title={Neural Window Fully-connected CRFs for Monocular Depth Estimation},
  author={Weihao Yuan and Xiaodong Gu and Zuozhuo Dai and Siyu Zhu and Ping Tan},
  booktitle={Conference on Computer Vision and Pattern Recognition (CVPR)},
  year={2022},
  url={https://api.semanticscholar.org/CorpusID:247223090}
}

@inproceedings{Chen2023Textto3DUG,
  title={Text-to-3D using Gaussian Splatting},
  author={Zilong Chen and Feng Wang and Huaping Liu},
  booktitle={Conference on Computer Vision and Pattern Recognition (CVPR)},
  year={2023},
  url={https://api.semanticscholar.org/CorpusID:263139613}
}

@inproceedings{liu2023zero,
  title={Zero-1-to-3: Zero-shot one image to 3d object},
  author={Liu, Ruoshi and Wu, Rundi and Van Hoorick, Basile and Tokmakov, Pavel and Zakharov, Sergey and Vondrick, Carl},
  booktitle={International Conference on Computer Vision (ICCV)},
  year={2023}
}

@inproceedings{tang2023dreamgaussian,
  title={Dreamgaussian: Generative gaussian splatting for efficient 3d content creation},
  author={Tang, Jiaxiang and Ren, Jiawei and Zhou, Hang and Liu, Ziwei and Zeng, Gang},
  booktitle={International Conference on Learning Representations (ICLR)},
  year={2023}
}

@inproceedings{Li2023OnSiteAF,
  title={On-Site Adaptation for Monocular Depth Estimation with a Static Camera},
  author={Huan Li and Matteo Poggi and Fabio Tosi and Stefano Mattoccia},
  booktitle={British Machine Vision Conference (BMVC)},
  year={2023},
  url={https://api.semanticscholar.org/CorpusID:267000624}
}

@article{Rasla2022TheRI,
  title={The Relative Importance of Depth Cues and Semantic Edges for Indoor Mobility Using Simulated Prosthetic Vision in Immersive Virtual Reality},
  author={Alex Rasla and Michael Beyeler},
  journal={ACM Symposium on Virtual Reality Software and Technology},
  year={2022},
  url={https://api.semanticscholar.org/CorpusID:251467956}
}

@inproceedings{Li2023TesttimeDA,
  title={Test-time Domain Adaptation for Monocular Depth Estimation},
  author={Zhi Li and Shaoshuai Shi and Bernt Schiele and Dengxin Dai},
  booktitle={International Conference on Robotics and Automation (ICRA)},
  year={2023},
  url={https://api.semanticscholar.org/CorpusID:259337418}
}

@inproceedings{metzer2022latentnerf,
  title={Latent-NeRF for Shape-Guided Generation of 3D Shapes and Textures},
  author={Metzer, Gal and Richardson, Elad and Patashnik, Or and Giryes, Raja and Cohen-Or, Daniel},
  booktitle={Conference on Computer Vision and Pattern Recognition (CVPR)},
  year={2023}
}

@inproceedings{white2006combining,
  title={Combining cues: Shape from shading and texture},
  author={White, Ryan and Forsyth, David A},
  booktitle={Conference on Computer Vision and Pattern Recognition (CVPR)},
  year={2006},
}

@inproceedings{schops2019bad,
  title={Bad slam: Bundle adjusted direct rgb-d slam},
  author={Schops, Thomas and Sattler, Torsten and Pollefeys, Marc},
  booktitle={Conference on Computer Vision and Pattern Recognition (CVPR)},
  year={2019}
}

@inproceedings{reizenstein21co3d,
	Author = {Reizenstein, Jeremy and Shapovalov, Roman and Henzler, Philipp and Sbordone, Luca and Labatut, Patrick and Novotny, David},
	Booktitle = {International Conference on Computer Vision (ICCV)},
	Title = {Common Objects in 3D: Large-Scale Learning and Evaluation of Real-life 3D Category Reconstruction},
	Year = {2021},
}

@inproceedings{MelasKyriazi2023RealFusion3R,
  title={RealFusion 360° Reconstruction of Any Object from a Single Image},
  author={Luke Melas-Kyriazi and Iro Laina and C. Rupprecht and Andrea Vedaldi},
  booktitle={Conference on Computer Vision and Pattern Recognition (CVPR)},
  year={2023},
  url={https://api.semanticscholar.org/CorpusID:261092262}
}

@inproceedings{Lorraine2023ATT3DAT,
  title={ATT3D: Amortized Text-to-3D Object Synthesis},
  author={Jonathan Lorraine and Kevin Xie and Xiaohui Zeng and Chen-Hsuan Lin and Towaki Takikawa and Nicholas Sharp and Tsung-Yi Lin and Ming-Yu Liu and Sanja Fidler and James Lucas},
  booktitle={International Conference on Computer Vision (ICCV)},
  year={2023},
  url={https://api.semanticscholar.org/CorpusID:259145353}
}

@inproceedings{Wang2018PseudoLiDARFV,
  title={Pseudo-LiDAR From Visual Depth Estimation: Bridging the Gap in 3D Object Detection for Autonomous Driving},
  author={Yan Wang and Wei-Lun Chao and Divyansh Garg and Bharath Hariharan and Mark E. Campbell and Kilian Q. Weinberger},
  booktitle={Conference on Computer Vision and Pattern Recognition (CVPR)},
  year={2018},
  url={https://api.semanticscholar.org/CorpusID:56177594}
}

@inproceedings{Park2024TestTA,
  title={Test- Time Adaptation for Depth Completion},
  author={Hyoungseob Park and Anjali W. Gupta and Alex Wong},
  booktitle={Conference on Computer Vision and Pattern Recognition (CVPR)},
  year={2024},
  url={https://api.semanticscholar.org/CorpusID:267411864}
}

@article{Ranftl2019TowardsRM,
  title={Towards Robust Monocular Depth Estimation: Mixing Datasets for Zero-Shot Cross-Dataset Transfer},
  author={Ren{\'e} Ranftl and Katrin Lasinger and David Hafner and Konrad Schindler and Vladlen Koltun},
  journal={Transactions on Pattern Analysis and Machine Intelligence},
  year={2019},
  volume={44},
  pages={1623-1637},
  url={https://api.semanticscholar.org/CorpusID:195776274}
}

@inproceedings{Wang2023ProlificDreamerHA,
  title={ProlificDreamer: High-Fidelity and Diverse Text-to-3D Generation with Variational Score Distillation},
  author={Zhengyi Wang and Cheng Lu and Yikai Wang and Fan Bao and Chongxuan Li and Hang Su and Jun Zhu},
  booktitle={Neural Information Processing Systems (NeurIPS)},
  year={2023},
  url={https://api.semanticscholar.org/CorpusID:258887357}
}

@inproceedings{Dosovitskiy2020AnII,
  title={An Image is Worth 16x16 Words: Transformers for Image Recognition at Scale},
  author={Alexey Dosovitskiy and Lucas Beyer and Alexander Kolesnikov and Dirk Weissenborn and Xiaohua Zhai and Thomas Unterthiner and Mostafa Dehghani and Matthias Minderer and Georg Heigold and Sylvain Gelly and Jakob Uszkoreit and Neil Houlsby},
  booktitle={International Conference on Learning Representations (ICLR)},
  year={2021},
  url={https://api.semanticscholar.org/CorpusID:225039882}
}

@article{Kerbl20233DGS,
  title={3D Gaussian Splatting for Real-Time Radiance Field Rendering},
  author={Bernhard Kerbl and Georgios Kopanas and Thomas Leimkuehler and George Drettakis},
  journal={ACM Transactions on Graphics (TOG)},
  year={2023},
  volume={42},
  pages={1 - 14},
  url={https://api.semanticscholar.org/CorpusID:259267917}
}

@article{Ye2024GauStudioAM,
  title={GauStudio: A Modular Framework for 3D Gaussian Splatting and Beyond},
  author={Chongjie Ye and Yinyu Nie and Jiahao Chang and Yuantao Chen and Yihao Zhi and Xiaoguang Han},
  journal={ArXiv},
  year={2024},
  url={https://api.semanticscholar.org/CorpusID:268732905}
}

@book{horn1989shape,
  title={Shape from Shading},
  author={Horn, Berthold K. P. and Brooks, Michael J.},
  year={1989},
  publisher={MIT Press},
  address={Cambridge, Massachusetts},
  isbn={978-0262081806}
}

@inproceedings{Xian2020StructureGuidedRL,
  title={Structure-Guided Ranking Loss for Single Image Depth Prediction},
  author={Xian, Ke and Zhang, Jianming and Wang, Oliver and Mai, Long and Lin, Zhe and Cao, Zhiguo},
  booktitle={Computer Vision and Pattern Recognition (CVPR)},
  year={2020},
  url={https://api.semanticscholar.org/CorpusID:219633501}
}

@inproceedings{Yin2019EnforcingGC,
  title={Enforcing Geometric Constraints of Virtual Normal for Depth Prediction},
  author={Wei Yin and Yifan Liu and Chunhua Shen and Youliang Yan},
  booktitle={International Conference on Computer Vision (ICCV)},
  year={2019},
  url={https://api.semanticscholar.org/CorpusID:198968133}
}

@article{zhang1999shape,
  title={Shape-from-shading: a survey},
  author={Zhang, Ruo and Tsai, Ping-Sing and Cryer, James Edwin and Shah, Mubarak},
  journal={IEEE Transactions on Pattern Analysis and Machine Intelligence},
  volume={21},
  number={8},
  pages={690--706},
  year={1999},
  publisher={IEEE},
  doi={10.1109/34.784284}
}

@inproceedings{mildenhall2020nerf,
 title={NeRF: Representing Scenes as Neural Radiance Fields for View Synthesis},
 author={Ben Mildenhall and Pratul P. Srinivasan and Matthew Tancik and Jonathan T. Barron and Ravi Ramamoorthi and Ren Ng},
 year={2020},
 booktitle={European Conference on Computer Vision (ECCV)},
}

@inproceedings{depthanything3,
  title={Depth Anything 3: recovering the visual space from any views},
  author={Haotong Lin and Sili Chen and Jun Hao Liew and Donny Y. Chen and Zhenyu Li and Guang Shi and Jiashi Feng and Bingyi Kang},
  year={2026},
  booktitle={International Conference on Learning Representations (ICLR)},
}

@inproceedings{blinn1977phong,
 title={Models of light reflection for computer synthesized pictures},
 author={James F. Blinn},
 year={1977},
 booktitle={Computer Graphics and Interactive Techniques},
}

@ARTICLE{759291,
  author={Ruo Zhang and Mubarak Shah},
  journal={IEEE Transactions on Systems, Man, and Cybernetics - Part A: Systems and Humans}, 
  title={Shape from intensity gradient}, 
  year={1999},
  volume={29},
  number={3},
  pages={318-325},
  doi={10.1109/3468.759291}}

@phdthesis{DBLP:phd/dnb/Scharr00,
  author       = {Hanno Scharr},
  title        = {Optimal operators in digital image processing},
  school       = {University of Heidelberg, Germany},
  year         = {2000},
  url          = {http://archiv.ub.uni-heidelberg.de/volltextserver/volltexte/2000/962/pdf/Diss.pdf},
  urn          = {urn:nbn:de:bsz:16-opus-9622},
  bibsource    = {dblp computer science bibliography, https://dblp.org}
}

@inbook{horn_brooks_variational, author = {Horn, Berthold K. P. and Brooks, Michael J.}, title = {The variational approach to shape from shading}, year = {1989}, isbn = {0262081830}, publisher = {MIT Press}, address = {Cambridge, MA, USA}, booktitle = {Shape from Shading}, pages = {173–214}, numpages = {42} }
}
\clearpage
\setcounter{page}{1}

\begin{strip}
\centering
\maketitlesupplementary

\includegraphics[width=\textwidth]{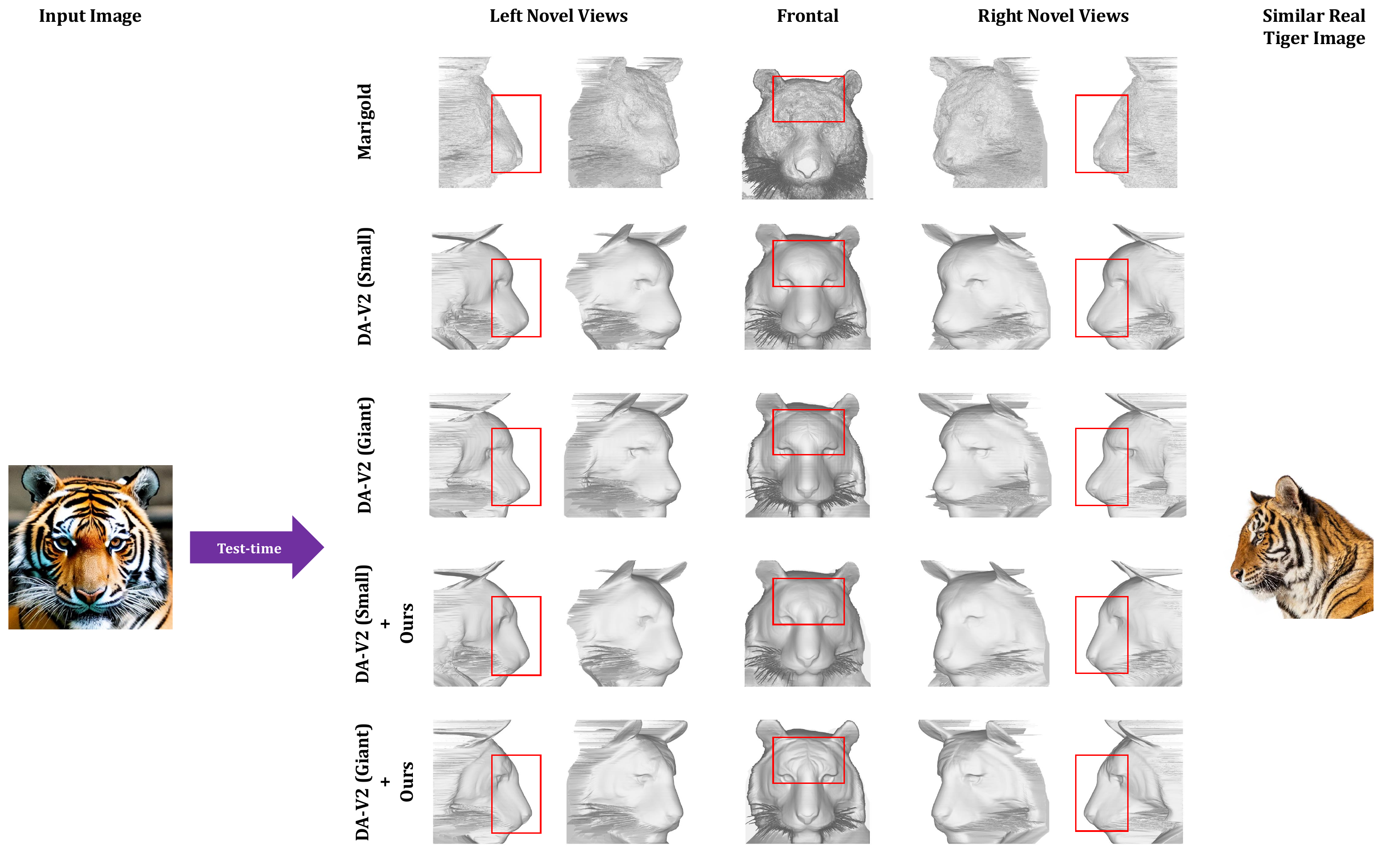}

\begin{minipage}{\textwidth}
\label{figsupp:teaser}
\captionof{figure}{\textbf{Biased predictions by baselines and our correction via re-lighting. Top:} Marigold and the larger DA-V2 variants struggle with the tiger image in ways similar to the small variant shown in the main paper teaser. \textbf{Bottom:} Our method applies to both the large and small variants of DA-V2, correcting the overall shape in each case and adding more detail when using the giant variant.}
\label{figsupp:teaser}
\end{minipage}
\end{strip}

\begin{figure*}
    \centering
    \includegraphics[width=\linewidth]{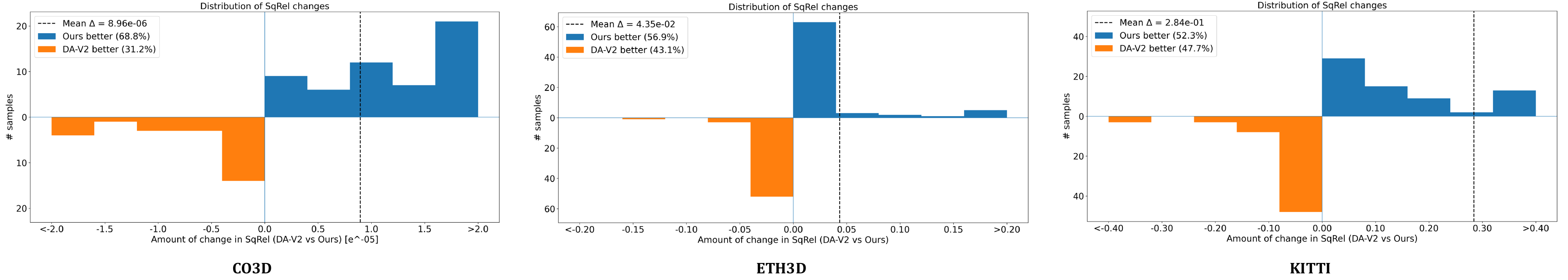}
    \caption{Histograms of per-sample change in SqRel with respect to DA-V2 on samples from CO3D, ETH3D, and KITTI. The x-axis shows $\Delta \text{SqRel} = \text{SqRel}_{\text{DA-V2}} - \text{SqRel}_{\text{Ours}}$ (for CO3D, scaled by $10^{-5}$ as indicated on the axis): blue bars on the right correspond to samples where our method achieves lower SqRel than DA-V2, and orange bars on the left correspond to samples where DA-V2 is better. The dashed vertical line marks the mean $\Delta \text{SqRel}$ for each dataset. On CO3D, the distribution is clearly skewed toward positive changes, with relatively few and smaller-magnitude failures. ETH3D and KITTI both exhibit a heavier positive tail. Overall, our method improves SqRel on more samples than it degrades, and the gains are larger in magnitude than the occasional losses, leading to a positive mean $\Delta \text{SqRel}$ on all three datasets.}
    \label{figsupp:improvementvsdegradation}
\end{figure*}

\begin{table*}[b!]
\caption{\textbf{Quantitative comparison with state-of-the-art monocular depth estimation methods.} DA-V2 and DA3 outperform the other baselines on CO3D and ETH3D, respectively, and our method further improves their performance. Therefore, we view our method as a self-supervised refinement strategy applicable to various base models. We compared different alignment metrics (see Sec.~\ref{supp:comparision}) and highlighted the comparable ones in gray and black, respectively. We advocate using the protocol marked in solid black.
}
\centering
\renewcommand{\arraystretch}{1.2}
\setlength{\tabcolsep}{5pt}
\resizebox{1.0\textwidth}{!}{
\begin{tabular}{l|lccc cccc ccccccc} 
\hline
\multirow{3}{*}{Dataset} 
 & \multirow{3}{*}{Method}
 & \multicolumn{3}{c}{Prediction} & 
 & \multicolumn{3}{c}{\textit{Higher is better} $\uparrow$} 
 &  & \multicolumn{6}{c}{\textit{Lower is better} $\downarrow$} \\ 
\cline{3-5} \cline{7-9} \cline{11-16}
 &  & Rel. Disp. & Rel. Depth & Abs. Depth &
 & $\delta_1$ & $\delta_2$ & $\delta_3$
 &  & AbsRel & RMSE & log10 & RMSE log & SI log & SqRel \\
\hline
\multirow{10}{*}{CO3D}
 &  $\text{Marigold}_{\text{ls-depth}}$ &  & $\checkmark$ &  &
 & 1.0 & 1.0 & 1.0
 &  & 0.00276 & 0.0685 & 0.001200 & 0.00366 & 0.366 & 0.000320 \\
 & \color{gray!50} $\text{Marigold}_{\text{ls-depth-disp}}$ &  & $\checkmark$ &  &
 & 1.0 & 1.0 & 1.0
 &  & 0.00275 & 0.0685 & 0.001196 & 0.00366 & 0.366 & 0.000320 \\
 & $\text{DepthPro}_{\text{ls-depth}}$ &  & & $\checkmark$  &
 & 1.0 & 1.0 & 1.0
 &  & 0.00242 & 0.0625 & 0.001053 & 0.00334 & 0.334 & 0.000286 \\
 & \color{gray!50} $\text{DepthPro}_{\text{ls-depth-disp}}$ &  & & $\checkmark$  &
 & 1.0 & 1.0 & 1.0
 &  & 0.00241 & 0.0625 & 0.001046 & 0.00334 & 0.334 & 0.000286 \\
\cline{2-16}
 & DA-V2$_{\text{ls-disp-depth}}$ & $\checkmark$ &  &  &
 & \textbf{1.0} & \textbf{1.0} & \textbf{1.0}
 &  & 0.00227 & 0.0602 & 0.000985 & 0.00321 & 0.321 & 0.000244 \\
 & \textbf{DA-V2 + Ours}$_{\text{ls-disp-depth}}$ & $\checkmark$ &  &  &
 & \textbf{1.0} & \textbf{1.0} & \textbf{1.0} 
 &  & \textbf{0.00223} & \textbf{0.0588} & \textbf{0.000968}
 &  \textbf{0.00314} & \textbf{0.314} & \textbf{0.000235} \\
 \cline{2-16}
 & DA3$_{\text{ls-depth}}$ &  & $\checkmark$ &  &
 & \textbf{1.0} & \textbf{1.0} & \textbf{1.0}
 &  & 0.00251 & 0.0667 & 0.00108 & 0.00357 & 0.357 & 0.000317 \\
 & \textbf{DA3 + Ours}$_{\text{ls-depth}}$ &  & $\checkmark$ &  &
 & \textbf{1.0} & \textbf{1.0} & \textbf{1.0}
 &  & \textbf{0.00238} & \textbf{0.0637} & \textbf{0.00103} & \textbf{0.00341} & \textbf{0.341} & \textbf{0.000294} \\
 \cline{2-16}
& \color{gray!50} DA-V2$_{\text{ls-disp}}$ & $\checkmark$ &  &  &
 & \textbf{1.0} & \textbf{1.0} & \textbf{1.0}
 &  & 0.00226 & 0.0602 & 0.000981 & 0.00321 & 0.321 & 0.000244 \\
& \color{gray!50} \textbf{DA-V2 + Ours}$_{\text{ls-disp}}$ & $\checkmark$ &  &  &
 & \textbf{1.0} & \textbf{1.0} & \textbf{1.0} 
 &  & \textbf{0.00222} & \textbf{0.0588} & \textbf{0.000964}
 &  \textbf{0.00314} & \textbf{0.314} & \textbf{0.000235} \\
\hline
\multirow{8}{*}{KITTI}
 & $\text{Marigold}_{\text{ls-depth}}$ &  & $\checkmark$ &  &
 & 0.889 & 0.978 & 0.992
 &  & 0.109 & 3.86 & 0.047 & 0.162 & 16.1 & 0.53 \\
& \color{gray!50} $\text{Marigold}_{\text{ls-depth-disp}}$ &  & $\checkmark$ &  &
 & 0.876 & 0.953 & 0.980
 &  & 0.110 & 5.63 & 0.051 & 0.181 & 17.9 & 0.84 \\
 & $\text{DepthPro}_{\text{ls-depth}}$ &  &  & $\checkmark$ &
 & 0.937 & 0.987 & 0.995
 &  & 0.086 & 2.74 & 0.037 & 0.132 & 13.0 & 0.30 \\
& \color{gray!50} $\text{DepthPro}_{\text{ls-depth-disp}}$ &  &  & $\checkmark$ &
 & 0.896 & 0.963 & 0.983
 &  & 0.096 & 6.65 & 0.045 & 0.186 & 18.3 & 3.21 \\
\cline{2-16}
 & DA-V2$_{\text{ls-disp-depth}}$ & $\checkmark$ &  &  &
 & 0.568 & 0.796 & 0.902
 &  & 0.305 & 7.01 & 0.118 & 0.348 & 33.6 & 2.49 \\
 & \textbf{DA-V2 + Ours}$_{\text{ls-disp-depth}}$ & $\checkmark$ &  &  &
 & \textbf{0.593} & \textbf{0.818} & \textbf{0.917}
 &  & \textbf{0.283} & \textbf{6.71} & \textbf{0.110} & \textbf{0.319} & \textbf{30.7} & \textbf{2.20} \\
 \cline{2-16}
& \color{gray!50} DA-V2$_{\text{ls-disp}}$ & $\checkmark$ &  &  &
 & 0.818 & 0.937 & 0.974
 &  & 0.323 & 130 & 0.062 & 0.256 & 25.4 & 1756 \\
& \color{gray!50} \textbf{DA-V2 + Ours}$_{\text{ls-disp}}$ & $\checkmark$ &  &  &
 & \textbf{0.823} & \textbf{0.940} & \textbf{0.976}
 &  & \textbf{0.276} & \textbf{105} & \textbf{0.060} & \textbf{0.243} & \textbf{24.1} & \textbf{1335} \\
\hline
\multirow{10}{*}{ETH3D}
 & $\text{Marigold}_{\text{ls-depth}}$ &  & $\checkmark$ &  &
 & 0.963 & 0.994 & 0.998
 &  & 0.058 & 0.476 & 0.0255 & 0.101 & 10.0 & 0.083 \\
& \color{gray!50} $\text{Marigold}_{\text{ls-depth-disp}}$ &  & $\checkmark$ &  &
 & 0.952 & 0.988 & 0.995
 &  & 0.072 & 4.666 & 0.0318 & 0.178 & 17.7 & 30.25 \\
 & $\text{DepthPro}_{\text{ls-depth}}$ &  &  & $\checkmark$ &
 & 0.966 & 0.993 & 0.997
 &  & 0.058 & 0.498 & 0.0237 & 0.077 & 7.69 & 0.158 \\
& \color{gray!50} $\text{DepthPro}_{\text{ls-depth-disp}}$ &  &  & $\checkmark$ &
 & 0.962 & 0.990 & 0.994
 &  & 0.065 & 5.489 & 0.0299 & 0.206 & 20.5 & 19.9 \\
\cline{2-16}
 & DA-V2$_{\text{ls-disp-depth}}$ & $\checkmark$ &  &  &
 & 0.884 & 0.956 & 0.978
 &  & 0.113 & 0.955 & 0.0448 & 0.153 & 15.1 & 0.391 \\
 & \textbf{DA-V2 + Ours}$_{\text{ls-disp-depth}}$ & $\checkmark$ &  &  &
 & \textbf{0.898} & \textbf{0.965} & \textbf{0.982}
 &  & \textbf{0.104} & \textbf{0.875} & \textbf{0.0413} & \textbf{0.143} & \textbf{14.1} & \textbf{0.347} \\
 \cline{2-16}
 & DA3$_{\text{ls-depth}}$ &  & $\checkmark$ &  &
 & \textbf{0.983} & \textbf{0.998} & \textbf{0.999}
 &  & 0.0461 & 0.373 & 0.0194 & 0.0631 & 6.27 & 0.100 \\
 & \textbf{DA3 + Ours}$_{\text{ls-depth}}$ &  & $\checkmark$ &  &
 & \textbf{0.983} & 0.997 & \textbf{0.999}
 &  & \textbf{0.0458} & \textbf{0.370} & \textbf{0.0193} & \textbf{0.0628} & \textbf{6.24} & \textbf{0.099} \\
 \cline{2-16}
& \color{gray!50} DA-V2$_{\text{ls-disp}}$ & $\checkmark$ &  &  &
 & \textbf{0.968} & 0.991 & 0.995
 &  & 0.198 & 35.69 & \textbf{0.0253} & 0.0998 & 9.93 & 1339 \\
 
& \color{gray!50} \textbf{DA-V2 + Ours}$_{\text{ls-disp}}$ & $\checkmark$ &  &  &
 & \textbf{0.968} & \textbf{0.992} & \textbf{0.996}
 &  & \textbf{0.148} & \textbf{23.27} & \textbf{0.0253} & \textbf{0.0941} & \textbf{9.36} & \textbf{850.9} \\
\hline
\end{tabular}
}
\label{tabsupp:quantitative}
\end{table*}

This supplemental document provides additional qualitative and quantitative comparisons, justifies the choice of DA-V2 as the baseline, and explains the challenge of comparing state-of-the-art depth estimation models due to mismatched depth representations and evaluation protocols.

\section{Additional Comparisons}

Fig.~\ref{figsupp:teaser} exemplifies how different depth estimation methods struggle with the tiger image from the main paper. The reconstructed dog-like shapes indicate a bias in the training data rather than an issue with the model architecture or depth representation. Our re-lighting is agnostic to training and corrects the bias from a dog-like to a tiger-like shape, regardless of whether the giant or small variant of DA-V2 is used. Quantitative improvements over other methods are listed in Table~\ref{tabsupp:quantitative} and discussed further below.

\begin{figure*}[b]
    \centering
    \includegraphics[width=\textwidth, trim={0cm 2.8cm 0cm 0cm},clip]{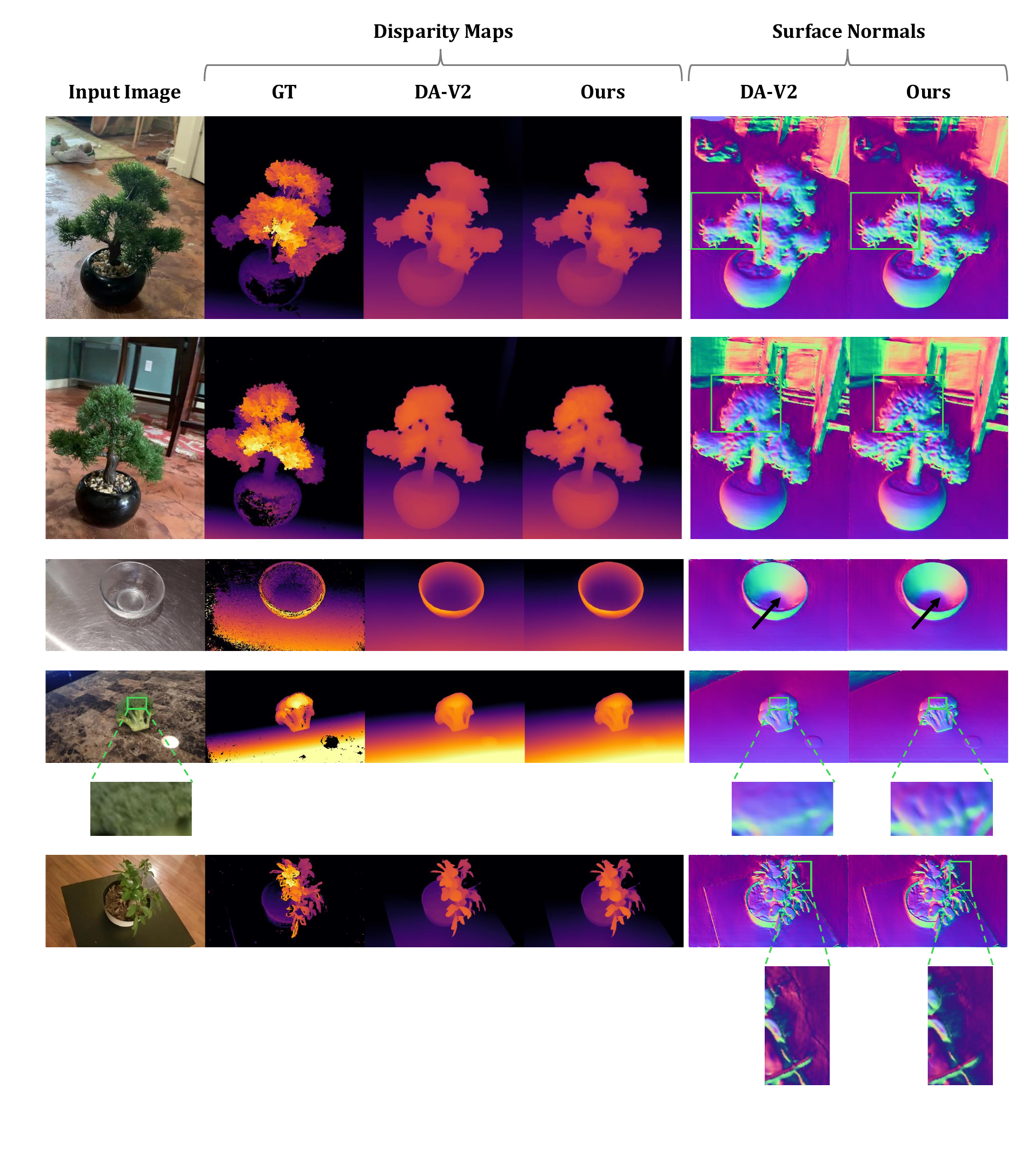}
    \caption{\textbf{Additional qualitative comparison on the CO3D dataset.}}
    \label{figsupp:co3d1}
    \clearpage
\end{figure*}

\begin{figure*}
    \centering
    \includegraphics[width=\textwidth, trim={0cm 2.5cm 0cm 0cm},clip]{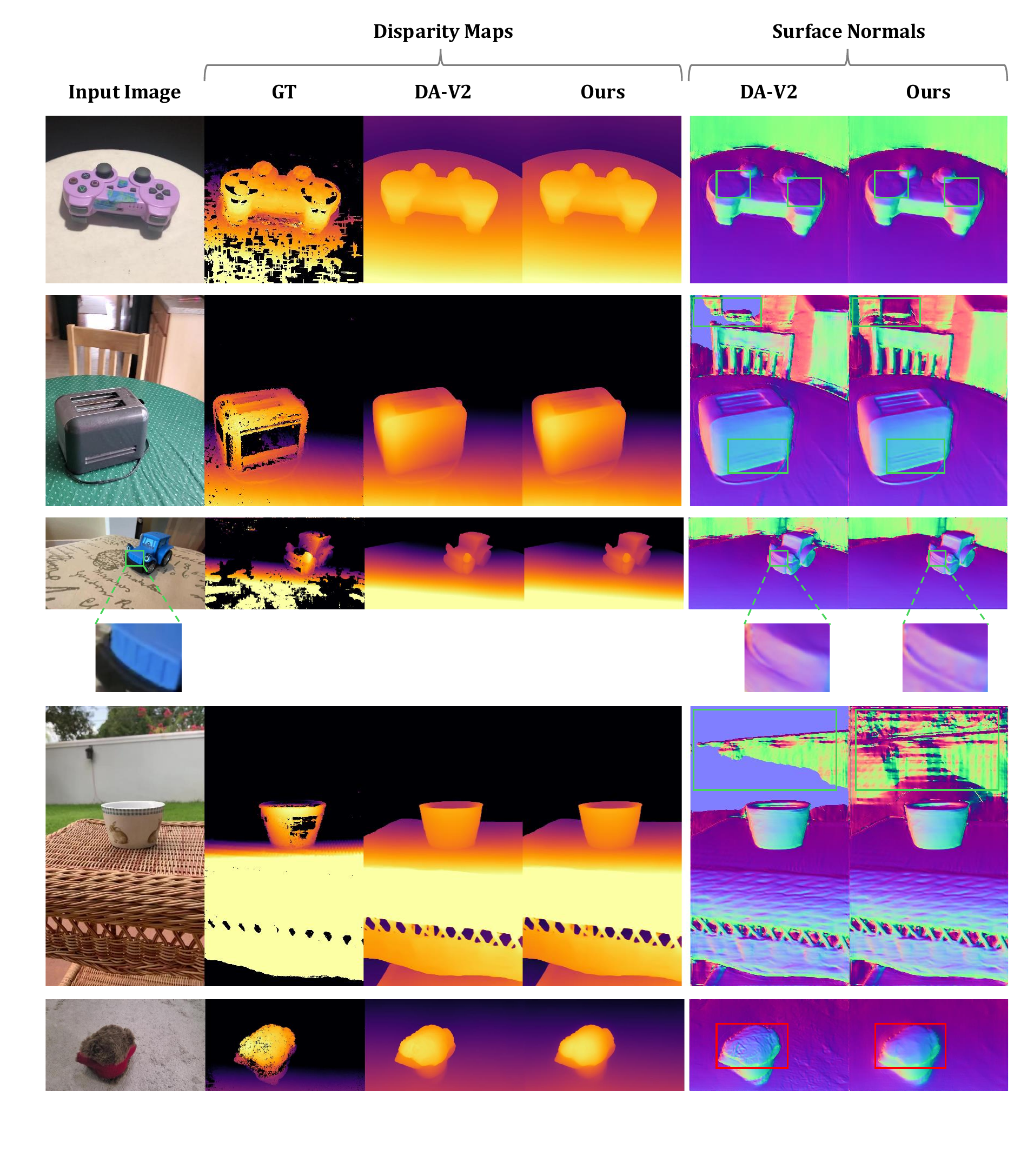}
    \caption{\textbf{Additional qualitative comparison on the CO3D dataset.} \textcolor{red}{Red} squares highlight occasional oversmoothing, a limitation of our method.}
    \label{figsupp:co3d2}
    \clearpage
\end{figure*}

\begin{figure*}
    \centering
    \includegraphics[width=\textwidth]{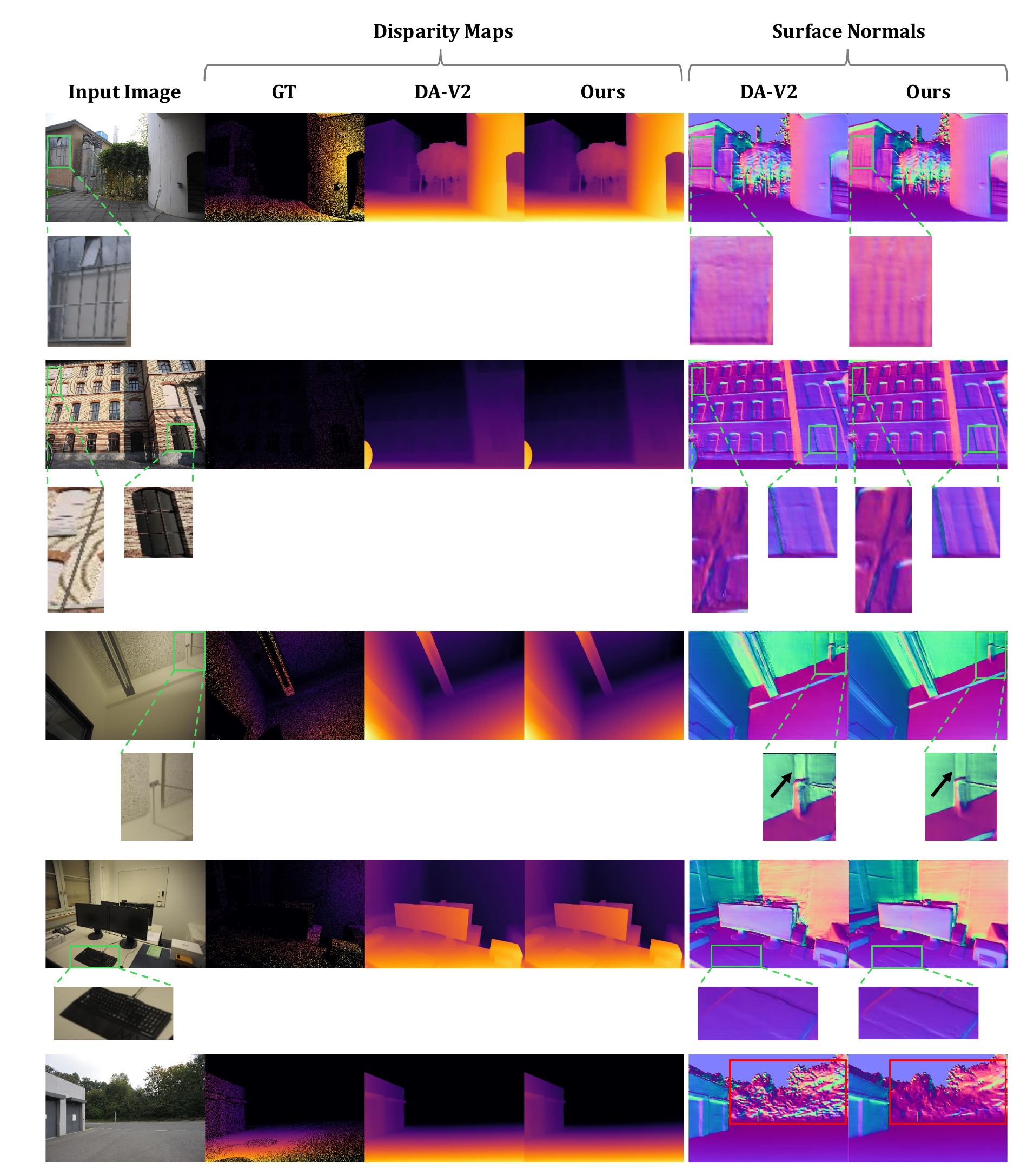}
    \caption{\textbf{Additional qualitative comparison on the ETH3D dataset.} \textcolor{red}{Red} squares highlight occasional oversmoothing, a limitation of our method.}
    \label{figsupp:eth3d}
    \clearpage
\end{figure*}

\begin{figure*}
    \centering
    \includegraphics[width=\textwidth, trim={0cm 2.5cm 0cm 0cm},clip]{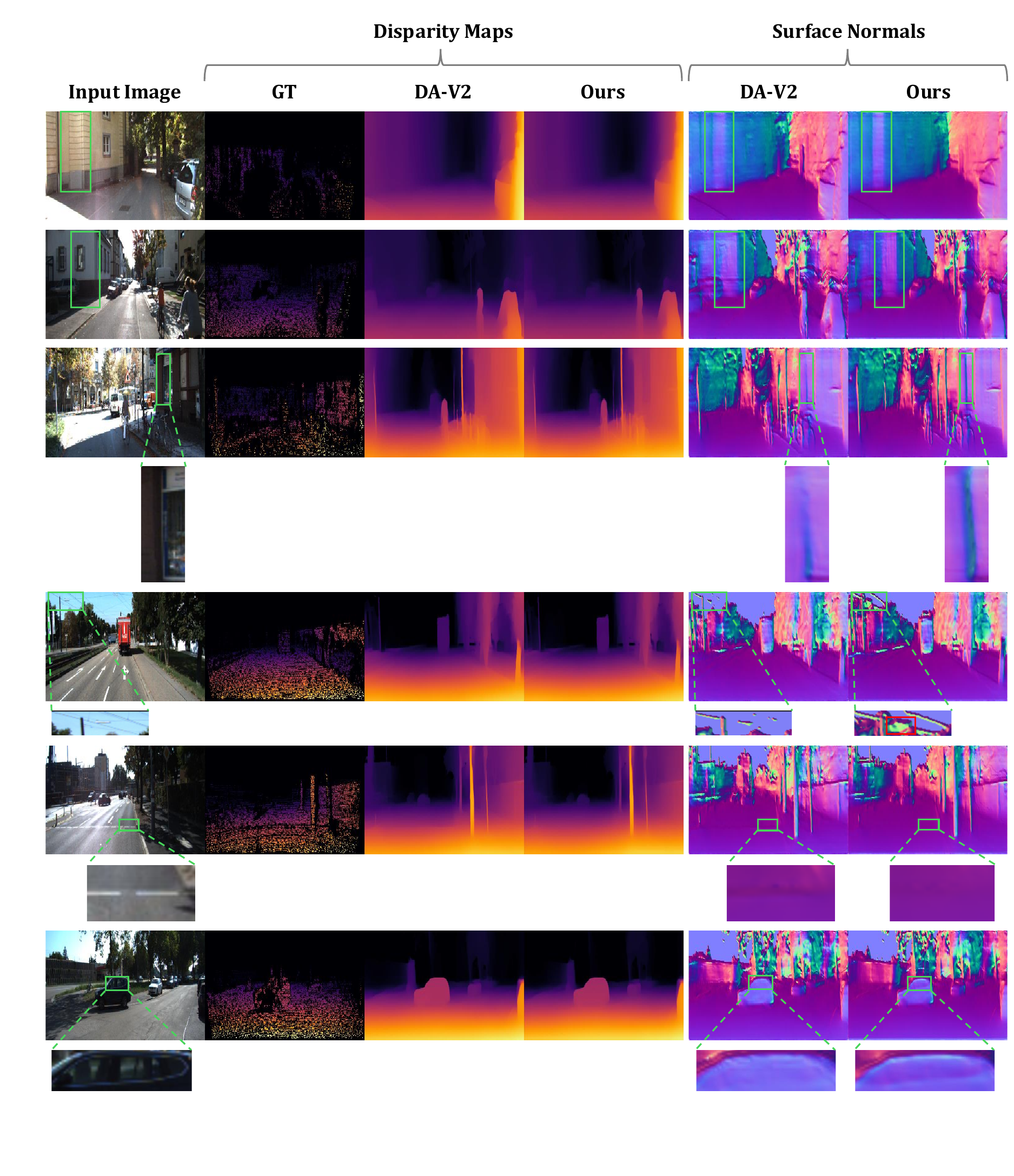}
    \caption{\textbf{Additional qualitative comparison on the KITTI dataset.} \textcolor{red}{Red} squares highlight occasional oversmoothing, a limitation of our method.}
    \label{figsupp:kitti}
    \clearpage
\end{figure*}

We present additional qualitative results in Figs.~\ref{figsupp:co3d1}, \ref{figsupp:co3d2}, \ref{figsupp:co3dda3}, \ref{figsupp:eth3d}, and~\ref{figsupp:kitti}. These figures also illustrate several remaining limitations of our method, highlighted by red boxes and discussed in the main document. Fig.~\ref{figsupp:improvementvsdegradation} further shows the per-sample change in SqRel relative to DA-V2, highlighting that our method delivers more frequent and larger-magnitude improvements across all three datasets. 

\section{Results with the DA-V2 Giant Backbone}

We used the ViT-S backbone for the main experiments because it performs almost as well as the larger ones while being significantly smaller (in terms of embedding size) and therefore more efficient to optimize. We also found that the giant model with the ViT-G encoder suffers from similar biases, such as the dog-like prediction for the tiger image in Fig.~\ref{figsupp:teaser}. Therefore, we used the smaller model for development and for the major comparisons.

At the time the paper was first written, the DA-V2 Giant weights were still available on Hugging Face via the following \href{https://huggingface.co/likeabruh/depth_anything_v2_vitg/tree/main}{link}, even though they had been removed from the official GitHub repository.

\section{Results with the DA3 Mono Large}
\begin{figure*}[b]
    \centering
    \includegraphics[width=\textwidth, trim={0cm 0cm 0cm 0cm},clip]{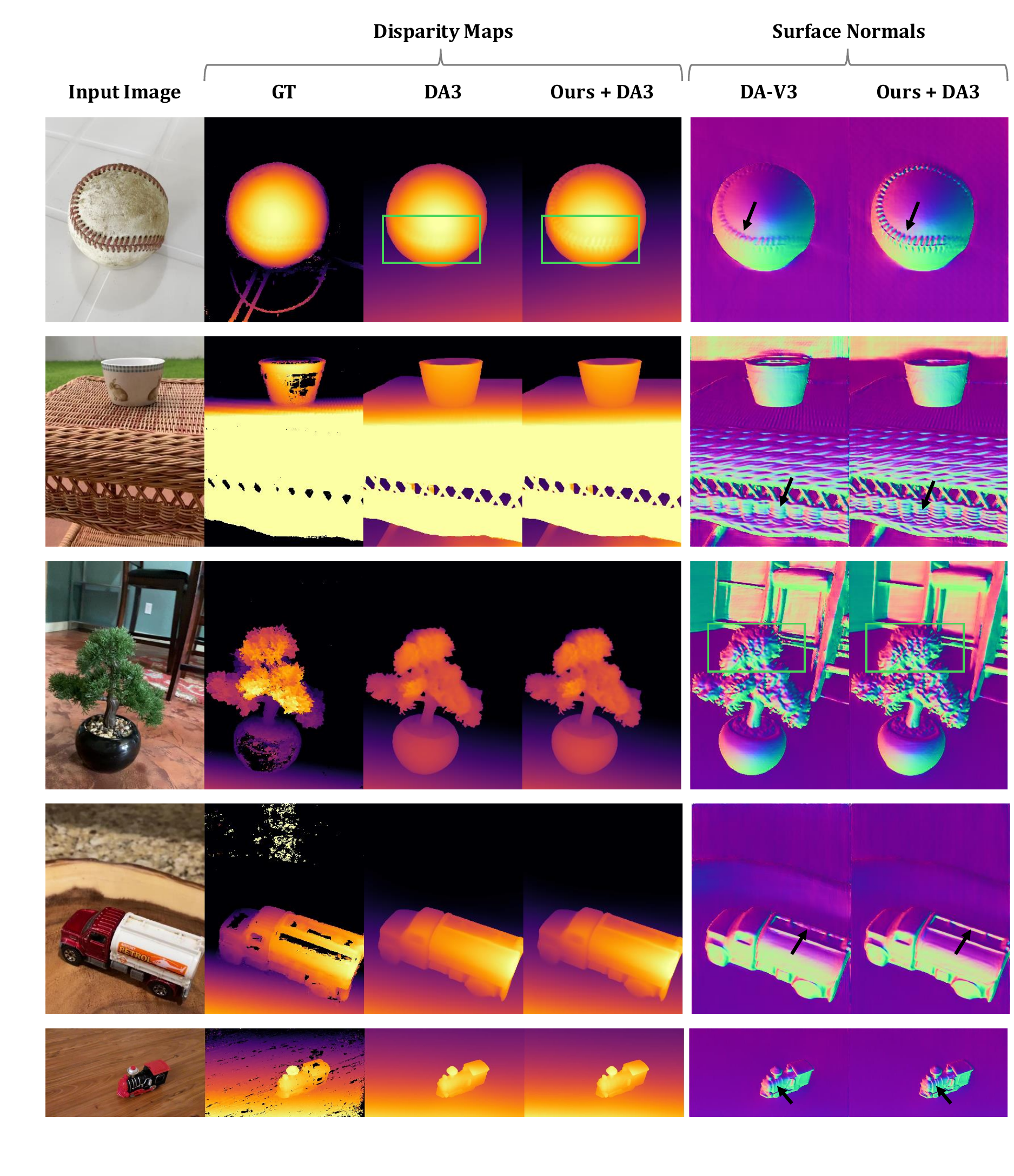}
    \caption{\textbf{Qualitative comparison against DA3 on the CO3D dataset.}}
    \label{figsupp:co3dda3}
    \clearpage
\end{figure*}

We evaluated the effectiveness of our refinement strategy on the state-of-the-art Depth Anything 3 (DA3)~\cite{depthanything3} (DA3MONO-LARGE), which predicts depth instead of disparity and employs a ViT-L encoder. Our method improves upon DA3 across both the CO3D and ETH3D datasets (see Table~\ref{tabsupp:quantitative}), setting a new state-of-the-art. Qualitative comparisons in Fig.~\ref{figsupp:co3dda3} further show that our approach significantly enhances details. Note that while the model predicts depth, we visualize the disparity maps for better contrast and to maintain visual consistency with other figures in this paper.

For DA3 experiments, we use a scaled orthographic camera, with its scale factor optimized from an initial value of 7.0. Additionally, we set a learning rate of $2 \times 10^{-4}$ for the embeddings and $1 \times 10^{-6}$ for the DPT weights. The regularization weight $\lambda_1$ is set to 1.0 for ETH3D's indoor/outdoor scenes and 10.0 for CO3D's close-up objects, respectively. Unless explicitly noted, all other hyperparameters are identical to those used in our DA-V2 experiments. Normal MSE in the main document is calculated from spatial depth gradients over pixels with valid neighboring depth values.

\section{Evaluation Protocol Details}
\label{supp:comparision}

Since monocular depth estimation is inherently scale-ambiguous, various depth representations and corresponding evaluation protocols have been proposed to support different invariances. Unfortunately, even among relative depth estimation methods, there are discrepancies in the evaluation protocols used. 

The classical approach is to align the relative depth prediction with the ground-truth depth in a least-squares sense before applying a range of metrics, which may include log transformations to reduce the impact of uncertainties. The benefit of aligning to the ground truth is that it provides interpretable results in metric space that are comparable to methods using absolute depth prediction.

DA-V2 deviated from this path by predicting disparity instead of relative depth. Hence, it was evaluated directly in disparity space using the same alignment procedure and metrics. However, this makes it incomparable to models that evaluate on depth.
An established alternative is to perform alignment in disparity space and then apply the metrics after converting disparity back to depth. However, least-squares fitting in disparity handles outliers very differently than when computed in depth. For instance, in disparity space, far estimates are less pronounced, leading to an unfair comparison with methods that perform alignment directly in depth space.

To provide an as-fair-as-possible comparison consistent with widely used evaluation protocols, we followed the procedure described in the main document: we first align to obtain absolute disparity and then perform a second alignment to minimize least-squares errors in the same space used by methods that output relative depth. We refer to this as least-squares disparity-and-depth (ls-disp-depth) alignment. To shed light on the effect of the different alignment methods, we compare them in Table~\ref{supp:comparision}.
The least-squares alignment performed directly in disparity space followed by depth conversion (ls-disp), as used in~\cite{ke2024repurposing}, performs better on CO3D than ls-disp-depth, but worse on the other two datasets. To further analyze this effect for methods predicting depth, we also mapped depth predictions into disparity space for a second alignment (ls-depth-disp) and observed the same trend. This experiment highlights the importance of the alignment procedure, and we conclude that the fairest comparison is to perform alignment in the same (depth) space, i.e., to use ls-disp-depth for methods operating on disparity and ls-depth for methods outputting depth directly. Note that the initial alignment in disparity space (ls-disp-depth) is inevitable for methods that output disparity, as it is required to obtain absolute disparity before converting disparity to depth.

\section{Baseline Selection}

To determine the suitability of DA-V2 as a baseline, we evaluated several existing methods (including Marigold, DepthPro, and DA-V2) on the CO3D dataset. We selected CO3D as our benchmark because its objects exhibit high detail and are relatively close to the camera (e.g., a toy truck instead of a real truck in KITTI). As a result, the depth estimates are more reliable, which aligns well with our goal of detail refinement, as motivated in the main document.

On CO3D, DA-V2 consistently outperformed all other methods. Marigold (a diffusion-based foundational model for depth estimation) captures coarse geometric structure; however, its reconstructed mesh, as shown in Fig.~\ref{figsupp:teaser}, exhibits significant high-frequency noise and artifacts. DepthPro provides an estimate of absolute scale, but it is often misled by toy versions of real objects (e.g., a toy truck), and even after normalizing for scale (and shift), it does not outperform DA-V2. This justifies our selection of DA-V2 as the baseline. At the time of the paper’s first draft, DA3 had not been publicly released. 

Notably, Table~\ref{tabsupp:quantitative} shows that DA-V2 performs better on CO3D but slightly worse on the other two datasets. This is consistent with the evaluations in DepthPro and Marigold, which report improvements over DA-V2 on these datasets. These results support the visual observation that DA-V2 excels at predicting fine details while lagging slightly in capturing overall scene composition and the relative scale of objects.

Another closely related work is BetterDepth~\cite{zhang2024betterdepth}, which, similar to our approach, focuses on refining the output of a pre-trained depth foundation model. However, its implementation is not publicly available.

\section{Ablation - Perspective Camera}
We conduct an ablation study on the camera model (Eq.~2 in the main paper) and the choice of $b=ms$ (Eq.~5 in the main paper). We compare a scaled orthographic camera, with its scale factor optimized from an initial value of 7.0, against a perspective camera, with its focal length jointly optimized from an initial value of 2.0. As shown in Table~\ref{tabsupp:cameraablation}, the scaled orthographic model with $b=0.1$ yields the lowest errors. We therefore adopt this configuration for all main experiments.

\begin{table}[t]
    \caption{\textbf{Ablation} of the camera model and $b$ parameter on the CO3D dataset.}
    \centering
    \renewcommand{\arraystretch}{1.2}
    \setlength{\tabcolsep}{5pt}
    \resizebox{0.99\linewidth}{!}{
    \begin{tabular}{lccccccc} 
    \hline
    Method & AbsRel $\downarrow$ & RMSE $\downarrow$ & log10 $\downarrow$ & RMSE log $\downarrow$ & SI log $\downarrow$ & SqRel $\downarrow$ \\ 
    \hline
    \textbf{Ours} ($\mK_\text{orth}$, $b=0.1)$ & \textbf{0.00223} & \textbf{0.0588} & \textbf{0.000968} & \textbf{0.00314} & \textbf{0.314} & \textbf{0.000235} \\
    \textbf{Ours-Perspective-Fixed} ($\mK_\text{persp}$, $b=0.1)$  & 0.00224 & 0.0591 & 0.000973 & 0.00315 & 0.315 & 0.000236\\
    \textbf{Ours-Perspective-Opt} ($\mK_\text{persp}$, $b_\text{init}=0.01$)  & 0.00225 & 0.0592 & 0.000976 & 0.00316 & 0.316 & 0.000237\\
    \hline
    \end{tabular}
    }
    \label{tabsupp:cameraablation}
\end{table}

\section{SfS Implementation Details}

We implement a simple shape from shading algorithm similar to \cite{horn_brooks_variational}.
We assume a Lambertian surface, 
with constant albedo and a single light at infinity such that the light direction $\mathbf{l}$ is unknown yet constant across all pixels.
We consider only brightness variations and therefore convert the input image to grayscale and 
set the incoming light intensity to $L_{\text{in}}=\max(\mI)$, where $\mI$ is the input image.

Under these assumptions, the image formation model reduces to a Lambertian dot product between the surface normal and the light direction:
\begin{equation*}
    \hat{\mI}(u,v) = \max\bigl(0,\; \mN(u,v)\cdot \mathbf{l}\bigr),
\end{equation*}
where $\hat{\mI}$ is the rendered image.

Motivated by the shape from intensity gradient technique~\cite{759291}, we initialize the normals using image gradients,
\begin{equation*}
    \mN(u,v) = \bigl(-\mI_u(u,v),\; -\mI_v(u,v),\; 1\bigr),
\end{equation*}
where $\mI_u$ and $\mI_v$ denote the partial derivatives of $\mI$ with respect to the spatial coordinates $u$ and $v$. 
We compute these derivatives using Scharr filters~\cite{DBLP:phd/dnb/Scharr00}.

To optimize the light direction and surface normals, we minimize a combination of a photometric loss and a smoothness loss, 
\begin{equation*}
    \mathcal{L} = \mathcal{L}_{\text{smooth}} + \lambda\,\mathcal{L}_{\text{photo}},
\end{equation*}
where $\lambda$ is a regularization parameter.  
The photometric loss minimizes the difference between the input and the rendered images,
\begin{equation*}
    \mathcal{L}_{\text{photo}}
    = \frac{1}{|\Omega|}
      \sum_{(u,v)\in\Omega}
      \bigl(\mI(u,v) - \hat{\mI}(u,v)\bigr)^2,
\end{equation*}
where $\Omega$ denotes the valid region defined by the object mask.
The smoothness loss penalizes spatial variation in the normals,
\begin{equation*}
    \mathcal{L}_{\text{smooth}}
    = \frac{1}{|\Omega|}
      \sum_{(u,v)\in\Omega}
      \left(
          \|\nabla \mN_u(u,v)\|_2^{2}
        + \|\nabla \mN_v(u,v)\|_2^{2}
      \right),
\end{equation*}
where $\mN_u$ and $\mN_v$ denote the first two components of the normal field.

This simple shading model closely matches the re-lighting procedure used in our main method, highlighting the benefit of our shading-based augmentation. 
Unlike full photometric reconstruction, which produces artifacts at texture boundaries or under complex real-world illumination, our re-lighting refinement succeeds with a simple and robust illumination model.


\end{document}